\documentclass{article}

\usepackage{arxiv}

\usepackage[utf8]{inputenc} % allow utf-8 input
\usepackage[T1]{fontenc}    % use 8-bit T1 fonts
\usepackage{url}            % simple URL typesetting
\usepackage{booktabs}       % professional-quality tables
\usepackage{amsfonts}       % blackboard math symbols
\usepackage{nicefrac}       % compact symbols for 1/2, etc.
\usepackage{microtype}      % microtypography
\usepackage[pdftex]{graphicx}
\usepackage{amssymb,amsfonts,amsmath,amscd,amsthm}
\usepackage{natbib}
\usepackage{mathrsfs}
\usepackage{rotating}
\usepackage{tikz-cd}
\usepackage{todonotes}
\usepackage{amsmath}
\usepackage[nolist]{acronym}
\usepackage{enumitem}
\usepackage[hidelinks]{hyperref}

\graphicspath{{figs/}}
\newcommand{\bbm}{\begin{bmatrix}}
\newcommand{\ebm}{\end{bmatrix}}
\newcommand{\beq}{\begin{equation}}
\newcommand{\eeq}{\end{equation}}
\newcommand{\bdis}{\begin{displaymath}}
\newcommand{\edis}{\end{displaymath}}
\newcommand{\beqn}[1]{\begin{subequations}\label{eq:#1}\begin{eqnarray}}
\newcommand{\eeqn}{\end{eqnarray}\end{subequations}}
\newcommand{\bfa}{{\bm{a}}}
\newcommand{\dbfa}{\dot{\bfa}}

\newcommand{\hbfa}{\hat{\bfa}}
\newcommand{\bfA}{\bm{A}}
\newcommand{\bfb}{\mathbf{b}}

\newcommand{\bff}{\bm{f}}
\newcommand{\hbff}{\hat{\bm{f}}}

\newcommand{\bfF}{\bm{F}}

\newcommand{\bfG}{\bm{G}}

\newcommand{\bfg}{\bm{g}}
\newcommand{\bfh}{\bm{h}}
\newcommand{\bfH}{\bm{H}}

\newcommand{\bfI}{\bm{I}}
\newcommand{\bfJ}{\bm{J}}

\newcommand{\bfp}{\bm{p}}
\newcommand{\hbfp}{\hat{\bm{p}}}
\newcommand{\dbfp}{\dot{\bfp}}

\newcommand{\bfP}{\bm{P}}
\newcommand{\hbfP}{\hat{\bm{P}}}

\newcommand{\bfq}{\bm{q}}

\newcommand{\bfQ}{\bm{Q}}

\newcommand{\bfr}{\bm{r}}
\newcommand{\hbfr}{\hat{\bm{r}}}

\newcommand{\bfR}{\bm{R}}

\newcommand{\dbfR}{\dot{\bfR}}
\newcommand{\hbfR}{\hat{\bfR}}

\newcommand{\bfS}{\bm{S}}

\newcommand{\bfT}{\bm{T}}
\newcommand{\hbfT}{\bm{\hat{T}}}

\newcommand{\bfu}{\mathbf{u}}

\newcommand{\bfv}{\bm{v}}

\newcommand{\hbfv}{\hat{\bfv}}

\newcommand{\dbfv}{\dot{\bfv}}
\newcommand{\bfW}{\bm{W}}
\newcommand{\bfw}{\bm{w}}

\newcommand{\hbfw}{\hat{\bm{w}}}
\newcommand{\bfx}{{\bm{x}}}

\newcommand{\hbfx}{\hat{\bm{x}}}

\newcommand{\bfy}{\bm{y}}

\newcommand{\boeta}{\bm{\eta}}

\newcommand{\boGamma}{\bm{\Gamma}}

\newcommand{\boomega}{\bm{\omega}}

\newcommand{\dboomega}{\dot{\boomega}}
\newcommand{\hboomega}{\hat{\boomega}}

\newcommand{\boPhi}{\bm{\Phi}}

\newcommand{\bosigma}{\bm{\sigma}}

\newcommand{\boxi}{\bm{\xi}}
\newcommand{\dboxi}{\dot{\bm{\xi}}}

\newcommand{\hboxi}{\mathbf{\hat{\boxi}}}

\newcommand{\bovartheta}{\bm{\vartheta}}
\newcommand{\bozeta}{\bm{\zeta}}

\newcommand{\bfzero}{\bm{0}}

\newcommand{\Exp}{\mathrm{Exp}}

\newcommand{\covR}{\bm{\mathcal{R}}}

\newcommand{\covP}{\bm{\mathcal{P}}}

\newcommand{\ned}{\{$n$\}}
\newcommand{\ecef}{\{$e$\}}
\newcommand{\body}{\{$b$\}}
\newcommand{\inertial}{\{$i$\}}
\newcommand{\geo}{\{$g$\}}

\usepackage{graphicx} % Required for inserting images
\usepackage[english]{babel}
\usepackage[utf8]{inputenc}
\usepackage{csquotes}% Recommended

\usepackage{amsmath}
\usepackage{amssymb}
\usepackage[per-mode = symbol]{siunitx}
\usepackage{xfrac}
\usepackage{bm}

\usepackage{tabularx}
\newcolumntype{Y}{>{\hsize=4\hsize}X}
\usepackage{booktabs}

\usepackage[capitalize]{cleveref}
\crefname{appendix}{App.}{Apps.}
\crefrangeformat{figure}{Figs.~#3#1#4--#5#2#6}
\crefrangeformat{table}{Tabs.~#3#1#4--#5#2#6}
\crefformat{equation}{(#2#1#3)}
\crefrangeformat{equation}{(#3#1#4)-(#5#2#6)}

\usepackage{todonotes}

\hypersetup{%
    pdftitle={},
    pdfauthor={},
    pdfkeywords={},
    pdfsubject={},
    pdfstartview=FitH,%
    bookmarksopen=true,%
    breaklinks=true,%
    colorlinks=true,%
    linkcolor=black,
    anchorcolor=black,%
    citecolor=black,
    filecolor=black,%
    menucolor=black,
    urlcolor=black
}%

\title{Derivations of Error-State Kalman Filter Kinematics for Globally Applicable Aided Inertial Navigation Systems}

\author{
 {\normalfont Antonia Hager$^{\dagger,\star}$ and Torleiv H. Bryne$^\star$} \\
 $^\dagger$\textit{Airbus Central Research and Technology, 
 Taufkirchen, Germany}  \\ 
 $^\star$\textit{Department of Engineering Cybernetics, 
 Norwegian University of Science and Technology, 
 Trondheim, Norway}
}

\date{}

\begin{document}
\begin{acronym}
\acro{NN}[NN]{Neural Networks}

\acro{TL}[TL]{Tolles-Lawson}

\acro{GNSS}[GNSS]{Global Navigation Satellite System}

\acro{MagNav}[MagNav]{Magnetic Anomaly Navigation}

\acro{EKF}[EKF]{extended Kalman filter}

\acro{ESKF}[ESKF]{error-state Kalman filter}

\acro{RMSE}[RMSE]{Root-Mean-Square Error}

\acro{CRLB}[CRLB]{Cramér-Rao Lower Bound}

\acro{IMA}[IMA]{Integrated Modular Avionics}

\acro{ML}[ML]{Machine-Learning}

\acro{IAE}[IAE]{Innovation-Based Adaptive Estimation}

\acro{IMU}[IMU]{inertial measurement unit}

\acro{RNP}[RNP]{Required Navigation Performance}

\acro{ECEF}[ECEF]{Earth-centered Earth-fixed}

\acro{FOGM}[FOGM]{First-Order Gauss-Markov}

\acro{IGRF}[IGRF]{International Geomagnetic Reference Field}

\acro{WGS84}[WGS84]{World Geodetic System 1984}

\acro{INS}[INS]{inertial navigation system}

\acro{RMSE}[RMSE]{Root Mean Square Error}

\acro{PNT}[PNT]{position, navigation and timing}

\acro{UAV}[UAV]{unmanned aerial vehicle}

\acro{MC}[MC]{Monte Carlo}
\end{acronym}

\setcounter{tocdepth}{1} 
\maketitle
\title{Derivations of Error-State Kalman Filter Kinematics for Globally Applicable Aided Inertial Navigation Systems}

\begin{abstract}
Global navigation systems require state estimation algorithms that handle Earth's curvature, Earth's rotation, and gravitational variations. These factors can typically be neglected in local navigation algorithms for robots, drones, etc.
In classical \acp{ESKF} the error state dynamics are trajectory-dependent. Invariant \acp{ESKF} utilize Lie Group symmetries to represent the error, which can render error propagation trajectory-independent for group-affine systems. Choosing between a standard filter (where position and velocity errors are defined additively in the navigation frame), a left-invariant filter (where errors are represented in the body frame) and a right-invariant filter (where errors are represented in the navigation/world frame) depends on system dynamics and sensor configuration.

This note presents the mathematical formulas for four classical and invariant \acp{ESKF} for globally applicable aided inertial navigation systems. It is intended to serve as a systematic reference for comparison and implementation.

\end{abstract}
\clearpage
\tableofcontents

\clearpage

\section{Introduction}
\label{sec:introduction}

Global and local navigation place different demands on the state estimation algorithm used. While in local navigation, speeds and distances are typically relatively small, and the use of a local Cartesian coordinate system that neglects Earth’s rotation is possible, precise global navigation requires consideration of Earth’s curvature, Earth’s rotation, and the inhomogeneity of the gravitational field. 
Furthermore, the higher speeds and distances typically encountered in global navigation of aircraft and spacecraft play a role in the numerical stability of the navigation algorithms, depending on the chosen coordinate system. Because classical error-state Kalman filters with an additive definition of the error rely on local linearization around the current state estimate, their error propagation and covariance dynamics are inherently trajectory-dependent.  
Invariant filters have advantageous properties for a group-affine system, since their error state dynamics are trajectory-independent when the invariance structure of the filter matches that of the system dynamics — specifically, a left-invariant filter achieves this for left-invariant dynamics, and a right-invariant filter for right-invariant dynamics. In the mismatched case, the error propagation retains a state dependence analogous to that of a classical EKF. For inertial navigation, where inputs are expressed in the body frame, the dynamics are left-invariant, making the left-invariant filter the natural choice for trajectory-independent error propagation. While world-frame measurements such as GNSS position are naturally compatible with a right-invariant error definition, the accuracy of covariance propagation between updates is typically more consequential for overall filter performance than the structure of the update step — particularly in low-update-rate or GPS-denied scenarios. The left-invariant filter is therefore generally preferred in practice.
A navigation problem is group-affine when its dynamics shares the symmetry of the local group-affine navigation problems. This leads to a better estimate of orientation in particular and therefore to improved navigation properties.
However, the assumption of a group-affine structure of the navigation problem fails when gravity cannot be assumed constant.

Here we systematically present the derivation of four global error-state Kalman filters formulations from the global navigation equations to enable their direct comparison: left- and right-invariant filters in global ECEF coordinates, based on the $SE_2(3)$ Lie group, in which position, velocity, and orientation errors (in the body and navigation coordinate systems, respectively) are naturally defined multiplicatively on the Lie group. In addition, a filter in ECEF coordinates with additive position and velocity errors, as well as a classical global navigation filter in geodetic coordinates, in which all three main state errors are defined additively in the navigation coordinate system. 

We also show the update equations and Jacobians for simple position measurements in the filter's respective navigation or world frame.

\subsection{Main contribution}

This note provides a unified, systematic reference for the mathematical continuous-time error-state dynamics of four global navigation \acp{ESKF} algorithms.

\subsection{Organization of the note}

The note starts with a preliminary \cref{sec:background} that establishes  
This note starts with a preliminary \cref{sec:background} that establishes

\begin{itemize}
    \item \cref{sec:notation}: The notation used throughout the note
    \item \cref{sec:navigation_equations}: The global navigation equations of a 15-state navigation system (position, velocity, attitude as well as accelerometer and gyroscope bias states) in geodetic/NED and in ECEF coordinates
    \item \cref{sec:lie_groups}: The basic Lie-Group formulations for Invariant Kalman filters
    \item \cref{sec:eskf}: The basic \acf{ESKF} equations.
\end{itemize}

Each of the following chapters is dedicated to one global navigation \ac{ESKF} formulation. The formulations differ in their underlying coordinate systems and error state definitions:

\begin{itemize}
    \item \cref{sec:geodetic_ESKF}: A standard geodetic/moving NED frame filter with position in geodetic coordinates, velocity in NED, and multiplicative attitude error.
    \item \cref{sec:LI_ECEF}: Left-Invariant (LI) \ac{ESKF} on the $SE_2(3)$ Lie-Group, with the nominal state in global ECEF coordinates and the error state in body frame.
    \item \cref{sec:RI_ECEF}: Right-Invariant (RI) \ac{ESKF} on the $SE_2(3)$ Lie-Group, with the nominal state in global ECEF coordinates and the error state in the ECEF navigation frame.
    \item \cref{sec:ECEF_ESKF}: A ``mixed'' ECEF closed-loop \ac{ESKF} with position and velocity errors defined additively in the navigation frame, and the body-frame attitude error on the $SO(3)$ Lie Group of rotations.
\end{itemize}

In each chapter, the full error-state kinematics of the respective system are derived, the system matrices are presented, and the correction mechanism is explained. Further, every chapter includes some brief comments on the mathematical and numerical properties of the filter formulation.

Finally, a brief tabular comparison of the four formulations is given in \cref{sec:comparison}, followed by the concluding remarks \cref{sec:conclusion}.

\clearpage
\section{Preliminaries}
\label{sec:background}

\subsection{Notation}
\label{sec:notation}
The notation used is as given in \Cref{tab:notation}.
\begin{table}[tb]
\begin{center}
\centering
\vspace{0.2cm}
\caption{\label{tab:notation}Notation}
%\small
\begin{tabularx}{\columnwidth}{>{\centering\arraybackslash}p{2cm} | X}
        \toprule \bfseries
        Symbol & \bfseries Description \\ 
        \midrule
        %\hline
        \inertial, \ecef, \body, \ned, \geo & Sub-/superscripts for inertial, ECEF-, body-, NED and geodetic coordinate frames  \\    
        \midrule
        $\bm p$, $\bm v$, $\bm{\vartheta}$ & Position, velocity, attitude\\
        $\mu, \lambda, h$ & Latitude, longitude, and altitude, $\mu \in [-\pi/2, \pi/2]$,   $\lambda \in (-\pi, \pi]$, $h\in \mathbb{R}^1$  \\
        $\phi$, $\theta$, $\psi$ & Roll, pitch, and yaw angles $\phi, \theta, \psi \in (-\pi, \pi]$ \\
        %\midrule
        $\bm{R}_{nb}$ & Rotation matrix from $\{b\}$ to $\{n\}$ frame\\
        $\bm{q}_{nb}$ & Unit quaternion from $\{b\}$ to $\{n\}$ frame\\
        \hline
        $R_0$ & Equatorial radius $R_0 = \SI{6378137.0}{\meter}$ \\
        $f$, $e$ & Flattening $f = 1 / 298.257223563$ and first eccentricity $e = [f(2-f)]^{1/2}$ of the reference ellipsoid \\
        $R_M$, $R_N$ & Earth's Meridial and Normal Radius \\
        $\omega_{ie}$ & Earth's rotation rate $\omega_{ie} = \SI{7.292115e-5 }{\radian\per\second}$\\
        %$\boomega^e_{ie}$  & Earth's rotation rate in ECEF frame, $\boomega^e_{ie} = \bfe^e_z\omega_{ie}$ \\
        $\bfg^n$ & Gravitational acceleration in NED frame
        \\
        \midrule
        $\bfa_m$, $\boomega_m$ & Acceleration and angular rate measured by IMU \\
        $\boeta_{\bfa}$, $\boeta_{\boomega}$ & Angular rate and acceleration measurement noise \\
        $\bfa_{\bfb}$, $\boomega_{\bfb}$ & Accelerometer and gyroscope bias\\%, sometimes also noted as $\hbfb_i^{\bfa}$ and $\hbfb_i^{\boomega}$\\
        $\boeta_{\bfa \bfb}$, $\boeta_{\boomega \bfb}$ & Accelerometer and gyroscope bias errors \\        
        \midrule
        $\bfx$, $\delta \bm x$ & Nominal state, Error state \\    
        $\bm{T}$%, $\delta \bm T$ 
        & 
        Extended pose on $SE_2(3)$%, corresponding error group 
        \\
%        $\boUpsilon$ & \textcolor{orange}{local increment (IMU integration)} \\
%        $\boGamma$ & \textcolor{orange}{global increment (Gravity integration)} \\
%        $\boxi$ & $SO(3)$ error state  \\
        $\boxi$ & Element of a Lie Group $\boxi = \begin{bmatrix}
            \boxi_1 & \boxi_2 & \boxi_3
        \end{bmatrix}^\top \in SE_2(3)$, $\boxi_1 \in SO(3)$ \\ 
        $\delta \bm p$, $\delta \bm v$ &  Position and velocity error vectors \\
        \midrule
        $\bfS(\cdot)$ & Skew-symmetric matrix \\
        $\Exp(\cdot)$ & Exponential map of $SO(3)$ or $SE_2(3)$ \\
        $\hbfR, \hbfT, \hboxi, \delta \hbfx, ...$ & Estimated quantities\\
        \bottomrule
    %\end{tabular}
    \end{tabularx}
\end{center}
\vspace{-0.7cm}
\end{table}

\subsection{Navigation equations}
\label{sec:navigation_equations}

We describe a system that includes 3D position, velocity, and attitude dynamics on rotating Earth, as well as accelerometer and gyroscope bias processes.

The estimated \ac{IMU} biases are always given in body frame, and their dynamics are described by the same Gauss-Markov process for all formulations:
\begin{subequations}
\begin{align}
    \dbfa^b_{\bfb} &= - p_{\bfa \bfb} \bfa^b_{\bfb}  + \boeta^b_{\bfa \bfb} \label{eq:d_ab}\\
    \dboomega^b_{\bfb} &= -p_{\boomega \bfb} \boomega^b_{\bfb} + \boeta^b_{\boomega \bfb}  \label{eq:d_omegab}
\end{align}
\label{eq:bias_cont}%
\end{subequations}
where $p_\bullet$ is the inverse of the Gass-Markov procees time constant. I.e $p_\bullet = 1/T_\bullet$.

\subsubsection{Geodetic/Local-Navigation-Frame formulation}

The most widespread global navigation equations describe the evolution of position in geodetic coordinates $\begin{bmatrix}\mu & \lambda &  h\end{bmatrix}^\top$, and velocity $\bfv^n_{eb} = [v_N, v_E, v_D]^\top$ and attitude $\bfR_{nb}$ (or
equivalently, the unit quaternion $\bfq_{nb}$) in a NED frame moving along with the vehicle \cite[Ch.~5.4]{Groves2013}, \cite[Ch.~3.7]{Titterton}, and are repeated here for reference.

The position kinematics are 
\begin{subequations}
\begin{align}
    \dot{\mu}     &= \frac{v_N}{R_M + h}, \label{eq:geo_mu}\\
    \dot{\lambda} &= \frac{v_E}{(R_N + h)\cos\mu}, \label{eq:geo_lam}\\
    \dot{h}       &= -v_D, 
    \label{eq:geo_h}
\end{align}
\label{eq:geo_pos_cont}
\end{subequations}
where $R_M$ and $R_N$ are the meridian and normal radii of curvature of
the WGS-84 ellipsoid evaluated at latitude $\mu$:

\begin{align}
    R_N &= \frac{R_0}{\sqrt{1 - e^2 \sin^2\mu}}, \quad
    R_M = \frac{R_0(1 - e^2)}{(1 - e^2 \sin^2\mu)^{1.5}}, 
\end{align}

The kinematics of the NED-frame velocity are
\begin{align}
    \dot{\bfv}^n_{eb}
    = \bfR_{nb}\bff^b_{ib}
      - (2\boomega^n_{ie} + \boomega^n_{en}) \times \bfv^n_{eb}
      + \bfg^n(\mu, h),
      \label{eq:geo_vel_cont}
\end{align}

where the Earth rate and transport rate in the navigation frame are
\begin{equation}
    \boomega^n_{ie} = \begin{bmatrix}\omega_{ie}\cos\mu \\ 0 \\ -\omega_{ie}\sin\mu\end{bmatrix},
    \qquad
    \boomega^n_{en} = \begin{bmatrix}
        \dfrac{v_E}{R_N+h} \\
        \dfrac{-v_N}{R_M+h} \\
        \dfrac{-v_E\tan\mu}{R_N+h}
    \end{bmatrix},
    \qquad
    \boomega^n_{in} = \boomega^n_{ie} + \boomega^n_{en},
    \label{eq:geo_rates}
\end{equation}
and $\bff^b_{ib} = \bff^b_m-\bfa^b_{\bfb}$ is the bias-corrected specific force in the
body frame.

Finally, the attitude kinematics of the body frame relative to the local NED frame are given by

\begin{align}
    \dbfR_{nb} &= -\bfS(\boomega^n_{in}) \bfR_{nb} + \bfR_{nb}\,\bfS(\boomega^b_{ib}), \notag\\
    &= \bfR_{nb}\,\bfS(\boomega^b_{nb}), \label{eq:geo_att_cont_R} \\
    \boomega^b_{nb} &= (\boomega_m - \boomega_\bfb)
                      - \bfR_{nb}^\top\boomega^n_{in},
    %\label{eq:geo_att_cont_R}
\end{align}
where $\boomega^b_m - \boomega^b_\bfb$ is the bias-corrected gyroscope output
and $\bfS(\cdot)$ denotes the skew-symmetric matrix operator.

Alternatively, this can be expressed using quaternions to avoid the polar singularity of Euler angle-based rotations.
The unit quaternion $\bfq_{nb} \in \mathbb{S}^3$ represents the same
rotation as $\bfR_{nb}$.  Its continuous-time kinematics are
\begin{align}
    \dot{\bfq}_{nb} = \tfrac{1}{2}\,\bfq_{nb} \otimes \bfq(\boomega^b_{nb}),
    \label{eq:geo_att_cont_q}
\end{align}
where $\bfq(\boomega) = [0,\; \boomega^\top]^\top$ is the pure
quaternion associated with $\boomega$, and $\otimes$ denotes the
Hamilton product.

\subsubsection{ECEF formulation}

The global navigation equations on rotating Earth expressed in ECEF coordinates read \cite[Ch.~5.3]{Groves2013}, \cite{Hager2025}
\begin{subequations}
\begin{align}
    \dbfR_{eb} &= - \bfS(\boomega^e_{ie}) \bfR_{eb} + \bfR_{eb} \bfS(\boomega^b_{ib})  \label{eq:dbfR2} \\
    \dbfv^e_{eb} &= - \bfS(2\boomega^e_{ie}) \bfv^e_{eb} + \bfR_{eb}\bff^b_{ib} + \bfg^e  - \bfS^2(\boomega^e_{ie}) \bfp^e_{eb}  \label{eq:dv}\\
    \dbfp^e_{eb} &= \bfv^e_{eb} \label{eq:dbfp2} %\\
    %
    %\dbfa^b_{\bfb} &= - p_{\bfa \bfb} \bfa^b_{\bfb}  + \boeta^b_{\bfa \bfb} \label{eq:d_ab}\\
    %
    %\dboomega^b_{\bfb} &= -p_{\boomega \bfb} \boomega^b_{\bfb} + \boeta^b_{\boomega \bfb}  \label{eq:d_omegab}
\end{align}
\label{eq:cont_inertial}%
\end{subequations}

The propagation of the nominal state in a discrete-time system is done by discretizing the system \eqref{eq:geo_pos_cont}--\eqref{eq:geo_att_cont_q} or \eqref{eq:cont_inertial}, and \eqref{eq:bias_cont} by a method of choice.

\subsection{Lie-Group Definitions}
\label{sec:lie_groups}

This section is based on \cite{Sola2021, Barfoot_2024, maurer_equivalence_2025} and re-states the most important Lie-Group definitions for Invariant Kalman Filter formulations.

\subsubsection{The $SO(3)$ Lie Group of Rotations}
The Lie Group of rotations is used to represent the orientation $\bfR \in SO(3)$ of a vehicle.
The exponential map of an element $\boxi_1 \in \mathbb{R}^3$ representing an incremental rotation vector $\boxi_1$ on the $SO(3)$ Lie Group of rotations is
\begin{align}
    \begin{split}
         \Exp(\boxi_1) &=  \exp (\bfS( \boxi_1)) \\
        &= 
        \left( \bfI_3 + \sin(\xi_1)\bfS(\bfu) + (1-\cos(\xi))\bfS^2(\bfu) \right)
        \in SO(3) \underset{\text{for\,small\,}\xi_1}{\approx} \left( \bfI_3 + \bfS(\boxi_1) \right)
    \end{split}
    \label{eq:exp_SO3}
\end{align}
where $\xi_1 = \| \boxi_1\|_2$ and $\bfu = \boxi_1/\xi_1$ and where \cref{eq:exp_SO3} can be recognized as the Rodrigues formula \cite{farrell_aided_2008,Groves2013,sola_quat} 

In quaternion form, the exponential map is given as
\begin{align}
    \bfq(\boxi_1) =
    \begin{bmatrix}
        \cos(\xi_1/2) \\
        \sin(\xi_1/2) \bfu
    \end{bmatrix}
    \underset{\xi_1 \ll 1}{\approx} 
    \begin{bmatrix}
        1\\
        \frac{1}{2}\boxi_1 \bf
    \end{bmatrix}.
\end{align}
as given in e.g. \cite{sola_quat}.

\paragraph{Jacobians of $SO(3)$.}

The \emph{right} and \emph{left} Jacobians of $SO(3)$ describe how the exponential
map responds to perturbations, and appear in the covariance reset step of the
invariant filters, defined by

\begin{align}
   \delta \bfx \leftarrow \gamma(\delta \bfx) = \delta \bfx \ominus \delta \hbfx.
   \label{eq:reset_fct}
\end{align}
%
%For $\boxi_1 \in \mathbb{R}^3$ with $\zeta = \|\boxi_1\|_2$:
The Jacobians are then given as
\begin{subequations}
\begin{align}
    \bfJ_r^{SO(3)}(\boxi_1)
        &= \bfI_3 - \frac{1-\cos\xi_1}{\xi_1^2}\bfS(\boxi_1)
                  + \frac{\xi_1-\sin\xi_1}{\xi_1^3}\bfS^2(\boxi_1),
    \label{eq:Jr_SO3} \\[4pt]
    \bfJ_l^{SO(3)}(\boxi_1)
        &= \bfA(\boxi_1)
         = \bfI_3 + \frac{1-\cos\xi_1}{\xi_1^2}\bfS(\boxi_1)
                  + \frac{\xi_1-\sin\xi_1}{\xi_1^3}\bfS^2(\boxi_1).
    \label{eq:Jl_SO3}
\end{align}
\label{eq:J_SO3}%
\end{subequations}
The two Jacobians are related by $\bfJ_r(\boxi_1) = \bfJ_l(-\boxi_1)$, i.e.\ they
differ only in the sign of the $\bfS(\boxi_1)$ term.
Their inverses are
\begin{subequations}
\begin{align}
    \bfJ_r^{SO(3),-1}(\boxi_1)
        &= \bfI_3 + \tfrac{1}{2}\bfS(\boxi_1)
          + \!\left(\frac{1}{\xi_1^2} - \frac{1+\cos\xi_1}{2\xi_1\sin\xi_1}\right)\!\bfS^2(\boxi_1),
    \label{eq:Jr_SO3_inv} \\[4pt]
    \bfJ_l^{SO(3),-1}(\boxi_1)
        &= \bfI_3 - \tfrac{1}{2}\bfS(\boxi_1)
          + \!\left(\frac{1}{\xi_1^2} - \frac{1+\cos\xi_1}{2\xi_1\sin\xi_1}\right)\!\bfS^2(\boxi_1).
    \label{eq:Jl_SO3_inv}
\end{align}
\label{eq:J_SO3_inv}%
\end{subequations}
For small $\xi_1$: $\bfJ_r \approx \bfI_3 - \tfrac{1}{2}\bfS(\boxi_1)$,\;
$\bfJ_l \approx \bfI_3 + \tfrac{1}{2}\bfS(\boxi_1)$.

\subsubsection{The $SE_2(3)$ Lie Group of Extended Poses}

The $SE_2(3)$ Lie Group of Extended Poses expands the representation of rotation on $SO(3)$ to the full 3D navigation state by adding velocity and position to the group element.

The exponential map of an element $\boxi = \begin{bmatrix}
    \boxi_1^\top & \boxi_2^\top & \boxi_3^\top
\end{bmatrix}^\top \in \mathbb{R}^9$ to the $SE_2(3)$ Lie Group of extended poses is
\begin{align}
    \begin{split}
        \Exp(\boxi) &= \exp\left( \boxi^\wedge \right)  \\
        &=\begin{bmatrix}
        \begin{array}{c|cc}
            \Exp(\boxi_1) & \bfA(\boxi_1)\boxi_2 & \bfA(\boxi_1)\boxi_3
            \\\midrule
            \bfzero_{2\times3} & \multicolumn{2}{c}{\bfI_2}
        \end{array}
        \end{bmatrix} \in SE_2(3),
    \end{split}
    \label{eq:Exp_SE23}
\end{align}
where
\begin{align}
    \boxi^\wedge &:=
    \begin{bmatrix}
        \bfS(\boxi_1) & \boxi_2 & \boxi_3 \\
        \bfzero_{1\times3} & 0 & 0 \\
        \bfzero_{1\times3} & 0 & 0
    \end{bmatrix} \in \mathbb{R}^{5\times5},
    \label{eq:hat_map}
\end{align}

Note that $\bfA(\boxi_1)$ is the \emph{left} Jacobian of $SO(3)$, $\bfA := \bfJ_l^{SO(3)}(\boxi_1)$,
and appears here as a structural property of the $SE_2(3)$ exponential map itself --
independent of whether a left- or right-invariant filter convention is later
chosen for the error definition. The filter convention instead determines which
group Jacobian, $\bfJ_r^{SE_2(3)}(\boxi)$ or
$\bfJ_r^{SE_2(3)}(-\boxi)$, is required for the covariance reset
(see \eqref{eq:cov_reset}).

The extended pose on the $SE_2(3)$ Lie-Group is defined as
\begin{align}
    \bfT &= \begin{bmatrix}\begin{array}{c|cc}\bfR &\bfv&\bfp\\\midrule\bfzero_{2\times3} & \multicolumn{2}{c}{\bfI_2} \end{array} \end{bmatrix} \in SE_2(3),
    \label{eq:se23_pose_def}
\end{align}
where $\bfR \in SO(3)$ is a rotation matrix mapping vectors from the body frame $\{b\}$
to the chosen reference frame $\{r\}$ (ECEF $\{e\}$ for the ECEF filters,
NED $\{n\}$ for the geodetic filter). The poses adjoint operator
\begin{align}
    \mathrm{ad}(\bfT) &= 
    \begin{bmatrix}
        \bfR  & \bfzero_{3 \times 3}& \bfzero_{3 \times 3}\\
        \bfS(\bfv)\bfR & \bfR & \bfzero_{3 \times 3}\\
        \bfS(\bfp)\bfR & \bfzero_{3 \times 3}& \bfR 
    \end{bmatrix}. \label{eq:ad_T}
\end{align}
as given by \cite{Barfoot_2024, Brossard2022}.

\paragraph{Right Jacobian of $SE_2(3)$.}

The right Jacobian of $SE_2(3)$ acts on $\boxi =
\begin{bmatrix}
    \boxi_1^\top & \boxi_2^\top & \boxi_3^\top
\end{bmatrix}^\top \in \mathbb{R}^9$ and is a $9\times9$ matrix.
It is defined through the adjoint algebra matrix with a structure similar to \cref{eq:ad_T}
\begin{align}
    \mathrm{ad}(\boxi) =
    \begin{bmatrix}
        \bfS(\boxi_1) & \bfzero_{3 \times 3}    & \bfzero_{3 \times 3}    \\
        \bfS(\boxi_2) & \bfS(\boxi_1)  & \bfzero_{3 \times 3}    \\
        \bfS(\boxi_3) & \bfzero_{3 \times 3}    & \bfS(\boxi_1)
    \end{bmatrix} \in \mathbb{R}^{9\times9},
    \label{eq:ad_SE23}
\end{align}
and given by the convergent series
\begin{align}
    \bfJ_r^{SE_2(3)}(\boxi)
    = \sum_{n=0}^{\infty} \frac{(-1)^n}{(n+1)!}\,\mathrm{ad}^n(\boxi).
    \label{eq:Jr_SE23_series}
\end{align}
Because $\bfS(\boxi_1)^3 = -\xi_1^2\bfS(\boxi_1)$ (Cayley--Hamilton), powers
$\mathrm{ad}^n$ for $n\geq5$ reduce to linear combinations of lower powers,
so the series closes exactly at fourth order:
\begin{align}
    \bfJ_r^{SE_2(3)}(\boxi)
    = \bfI_9 - c_1\,\mathrm{ad}
             + c_2\,\mathrm{ad}^2
             - c_3\,\mathrm{ad}^3
             + c_4\,\mathrm{ad}^4,
    \label{eq:Jr_SE23}
\end{align}
with coefficients ($\xi_1 = \|\boxi_1\|_2$, $s = \sin\xi_1$, $c = \cos\xi_1$):
\begin{subequations}
\begin{align}
    c_1 &= \frac{4 - \xi_1 s - 4c}{2\xi_1^2}, &
    c_2 &= \frac{4\xi_1 - 5s + \xi_1 c}{2\xi_1^3}, \\[4pt]
    c_3 &= \frac{2 - \xi_1 s - 2c}{2\xi_1^4}, &
    c_4 &= \frac{2\xi_1 - 3s + \xi_1 c}{2\xi_1^5}.
\end{align}
\label{eq:Jr_SE23_coeffs}%
\end{subequations}
For $\xi_1 < 10^{-2}$, \eqref{eq:Jr_SE23_series} is used directly and
converges in 5--6 terms since $\|\mathrm{ad}\| \sim \xi_1$.

The covariance reset in the LI filter uses $\bfJ_r^{SE_2(3)}(\boxi)$,
while the RI filter uses $\bfJ_r^{SE_2(3)}(-\boxi)$.
The full reset Jacobian is
\begin{align}
    \boGamma = \begin{bmatrix}
        \bfJ_r^{SE_2(3)}(\boxi) & \bfzero_{9\times6} \\
        \bfzero_{6\times9} & \bfI_6
    \end{bmatrix},
    \qquad
    \hbfP_k = \boGamma \bfP_{k|k} \boGamma^\top,
    \label{eq:cov_reset}
\end{align}
where the $\bfI_6$ block acts on the bias error states, which live in flat
$\mathbb{R}^6$ and require no curvature correction.

\subsection{Error-State Kalman Filter}
\label{sec:eskf}

The true system state $\bfx$ is a composition of the nominal state estimate and an error:
\begin{align}
    \bfx_\text{true} = \hbfx_\text{nom} \oplus \delta \bfx
    \label{eq:system_state}
\end{align}

Instead of estimating $\bfx$ directly using the dynamics of the full navigation state, in the \ac{ESKF} approach the filter models the dynamics of the error

\begin{align}
    \delta \bfx = \bfx_\text{true} \ominus \hbfx_\text{nom}
\end{align}

and uses the estimated error $\delta \hbfx$ to obtain an estimate of the nominal system state $\hbfx_\text{nom} := \hbfx$ 
The continuous-time error dynamics,

\begin{align}
    \delta \dot{\bfx} = \bfF \delta \bfx + \bfG \bfw
\end{align}

where $\bfF$ is the continuous-time system or dynamics matrix, $\hbfw$ is the noise, and $\bfG$ is the continuous-time noise coupling matrix. are derived from inserting \eqref{eq:system_state} into the chosen nominal state dynamic model (i.e., the global inertial navigation equations) and solving the (linearized) equations for $\delta \bfx$.

The general Kalman Filter prediction step over a discretization time interval $\Delta t := t_k - t_{k-1}$ is then:
\begin{align}
    \delta \hbfx _{k|k-1} &= \boPhi_{k-1} \cdot \delta \hbfx_{k-1} \label{eq:dx_pred}\\
    \bfP_k &= \boPhi_{k-1} \bfP_{k-1} \boPhi_{k-1}^\top + \bfG_d \bfQ \bfG_d^\top \notag \\
    \bfP_k &\triangleq \boPhi_{k-1} \bfP_{k-1} \boPhi_{k-1}^\top + \bfQ_d . 
    \label{eq:P_pred}
\end{align}
where $\boPhi$ is the discretized dynamic matrix of the system, $\bfG_d$ the discretized noise input matrix, and $\bm \bfQ_d$ is the discrete time process covariance matrix. These discrete-time matrices are obtained from the continuous-time error state dynamics matrices $\bfF$ and $\bfG$ by a discretization method of choice. However, for closed-loop or feedback ESKF the \cref{eq:dx_pred} is not necessary since the error state is always reset to zero after every measurement update.

The standard \ac{ESKF} update equations to correct the error state based on the measurements is 
\begin{align}
    \bfW_k &= \covP_{k|k-1} \bfH_k^\top (
    \bfH_k \covP_{k|k-1} \bfH^\top_k + \covR_{k}
    )^{-1} \\
    \delta \hbfx_k &= \bfW_k ( \bfy_k - \bfh(\hbfx_{k|k-1}) ) \\
    \covP_{k} &= (\bfI_{17} - \bfW_k \bfH_k) \covP_{k|k-1}.
\end{align}

\clearpage
\section{Geodetic-/North-East-Down-Frame ESKF}
\label{sec:geodetic_ESKF}

This section repeats the standard error state kinematic equations derived from the full global inertial navigation equations in \eqref{eq:geo_pos_cont}--\eqref{eq:geo_att_cont_R} as in, e.g., \cite{Titterton}, and the Jacobian for a simple position update in geodetic coordinates.

\subsection{Error Definition}

The geodetic ESKF expresses all errors additively in the local NED frame.
The nominal state is
\begin{equation}
    \bfx = \begin{bmatrix}
        \bfp^{g\top} &
        \bfv_{nb}^{n\top} &
        \bovartheta_{nb}^{n\top} & 
        \bfa_\bfb^{b\top} & \boomega_\bfb^{b\top}
    \end{bmatrix}^\top \in \mathbb{R}^{15},
\end{equation}
where $\bfp^g = \begin{bmatrix}\mu & \lambda & h\end{bmatrix}^\top$ is the geodetic position, $\bovartheta = \begin{bmatrix}    \phi & \theta & \psi\end{bmatrix}^\top$ are the roll, pitch, and yaw angles parametrising the rotation matrix $\bfR_{nb}$ (or equivalently the
quaternion $\bfq_{nb}$), and $\bfa_\bfb$, $\boomega_\bfb$ are the
IMU biases in the body frame.

Position and velocity errors are defined additively:
\begin{subequations}
\begin{alignat}{3}
    \delta\bfv_{nb}^n
        &:= \bfv_{nb}^n - \hat{\bfv}_{nb}^n,
        \label{eq:geo_vel_err} \\
    \delta\bfp
        &:= \begin{bmatrix}\delta\mu \\ \delta\lambda \\ \delta h\end{bmatrix}
         := \begin{bmatrix}
                \mu - \hat{\mu} \\
                \lambda - \hat{\lambda} \\
                h - \hat{h}
            \end{bmatrix}.
        \label{eq:geo_pos_err}
\end{alignat}
\label{eq:geo_errors}%
\end{subequations}

The true attitude is related to the nominal through a left-multiplicative
correction on $SO(3)$:
\begin{align}
    \bfR_{nb} = \exp(\delta\bovartheta^n_{nb})\,\hat{\bfR}_{nb},
    \quad
    \bfq_{nb} = \bfq(\delta\bovartheta^n_{nb}) \otimes \hat{\bfq}_{nb},
    \label{eq:geo_att_err_exact}
\end{align}
where $\delta\bovartheta^n_{nb} \in \mathbb{R}^3$ is the navigation-frame tilt
error vector and $\exp(\cdot)$ is the exponential map on $SO(3)$
\eqref{eq:exp_SO3}.
 
For the linearised error-state dynamics, \eqref{eq:geo_att_err_exact}
is approximated to first order in $\delta\bovartheta^n_{nb}$:
\begin{align}
    \bfR_{nb}
    \approx (\bfI_3 - \bfS(\delta\bovartheta^n_{nb}))\hat{\bfR}_{nb}.
    \label{eq:geo_att_err}
\end{align}
This is the standard linearisation used in all ESKF formulations. It is
valid as long as $\|\delta\bovartheta^n_{nb}\|$ remains small, which is
guaranteed by the filter reset after each update.

The full error state is:
\begin{equation}
   \delta \bfx = \begin{bmatrix}
       \delta\bfp^{g\top} &
       \delta\bfv_{nb}^{n\top} &
       \delta\bovartheta_{nb}^{n\top} &
       \delta\bfa_\bfb^{b\top} &
       \delta\boomega_\bfb^{b\top}
   \end{bmatrix}^\top \in \mathbb{R}^{15}.
   \label{eq:geo_error_state}
\end{equation}

\subsection{Error State Kinematics}
\label{sec:geo_err_kinematics}

\subsubsection{Continuous-time Error State kinematics}

The linearized error state kinematics in geodetic/local coordinates are standard in inertial navigation system technology and sometimes referred to as the \textit{Pinson Error State Model}. Their derivation from linearizing \eqref{eq:geo_pos_cont}--\eqref{eq:geo_att_cont_R} can, e.g., be found in \cite[Ch.~12.3]{Groves2013} and is not repeated here. We present the resulting equations for reference:

\begin{subequations}
\begin{align}
\begin{split}
\delta\dot{\bfp}^g
    &= \bfF_{pp}\,\delta\bfp^g + \bfF_{pv}\,\delta\bfv^n_{nb}
\end{split}\\
\begin{split}
\delta\dot{\bfv}_{nb}^n
    &= \bfF_{vp}\,\delta\bfp^g + \bfF_{vv}\,\delta\bfv^n_{nb}
       - \bfS(\bfR_{nb}\bff_{ib}^b)\,\delta\bovartheta^n_{nb}
       + \bfR_{nb}\,\delta\bfa_\bfb^b
       + \bfR_{nb}\,\boeta_{\bfa\bfb}
\end{split}\\
\begin{split}
\dot{\delta\bovartheta^n_{nb}}
    &= \bfF_{\vartheta p}\,\delta\bfp^g + \bfF_{\vartheta v}\,\delta\bfv^n_{nb}
       + \bfF_{\vartheta \vartheta }\,\delta\bovartheta^n_{nb}
       - \bfR_{nb}\,\delta\boomega_\bfb^b
       - \bfR_{nb}\,\boeta_{\boomega\bfb}
\end{split}\\
\begin{split}
\delta\dot{\bfa}_\bfb^b
    &= -p_{\bfa\bfb}\,\delta\bfa_\bfb^b + \boeta_{\bfa\bfb}^b
\end{split}\\
\begin{split}
\delta\dot{\boomega}_\bfb^b
    &= -p_{\boomega\bfb}\,\delta\boomega_\bfb^b + \boeta_{\boomega\bfb}^b
\end{split}
\end{align}
\label{eq:geo_err_kinematics}%
\end{subequations}

The attitude coupling $-\bfS(\bfR_{nb}\bff_{ib}^b)\delta\bovartheta^n_{nb}$
arises from the linearisation of the true rotation
$\bfR_{nb} \approx (\bfI-\bfS(\delta\bovartheta^n_{nb}))\hat{\bfR}_{nb}$.
The term $\bfF_{\vartheta \vartheta }\,\delta\bovartheta^n_{nb} = -\bfS(\boomega^n_{in})\delta\bovartheta^n_{nb}$
transports the tilt error in the rotating navigation frame.

\subsubsection{System Matrix}

The position sub-blocks $\bfF_{pp}$, $\bfF_{pv}$ from \eqref{eq:geo_pos_cont} are
\begin{equation}
    \bfF_{pp} =
    \begin{bmatrix}
        0 & 0 & -\dfrac{v_N}{(R_M + h)^2} \\[8pt]
        \dfrac{v_E\tan\mu}{(R_N + h)\cos\mu} & 0 & -\dfrac{v_E}{(R_N + h)^2\cos\mu} \\[8pt]
        0 & 0 & 0
    \end{bmatrix},
    \qquad
    \bfF_{pv} = \mathrm{diag}\!\left(
        \frac{1}{(R_M + h)},\;
        \frac{1}{(R_N + h)\cos\mu},\;
        -1
    \right).
    \label{eq:geo_App}
\end{equation}

The velocity sub-blocks from \eqref{eq:geo_vel_cont} are
\begin{equation}
    \bfF_{vp} =
    \begin{bmatrix}
        -v_E\!\left(2\omega_{ie}\cos\mu + \dfrac{v_E}{(R_N + h)\cos^2\mu}\right)
        & 0
        & \dfrac{v_E^2\tan\mu - v_Nv_D}{(R_M + h)^2} \\[10pt]
        2\omega_{ie}(v_N\cos\mu - v_D\sin\mu) + \dfrac{v_Nv_E}{(R_N + h)\cos^2\mu}
        & 0
        & -v_E\dfrac{v_N\tan\mu + v_D}{(R_N + h)^2} \\[10pt]
        2\omega_{ie}v_E\sin\mu
        & 0
        & \dfrac{v_N^2}{(R_M + h)^2} + \dfrac{v_E^2}{(R_N + h)^2}
    \end{bmatrix},
    \label{eq:geo_Avp}
\end{equation}
\begin{equation}
    \bfF_{vv} =
    \begin{bmatrix}
        \dfrac{v_D}{(R_M + h)}
        & -2\!\left(\omega_{ie}\sin\mu + \dfrac{v_E\tan\mu}{(R_N + h)}\right)
        & \dfrac{v_N}{(R_M + h)} \\[10pt]
        2\omega_{ie}\sin\mu + \dfrac{v_E\tan\mu}{(R_N + h)}
        & \dfrac{v_N\tan\mu + v_D}{(R_N + h)}
        & 2\omega_{ie}\cos\mu + \dfrac{v_E}{(R_N + h)} \\[10pt]
        -\dfrac{2v_N}{(R_M + h)}
        & -2\!\left(\omega_{ie}\cos\mu + \dfrac{v_E}{(R_N + h)}\right)
        & 0
    \end{bmatrix}.
    \label{eq:geo_Avv}
\end{equation}

The attitude sub-blocks from \eqref{eq:geo_att_cont_R} are
\begin{align}
    \bfF_{\vartheta p} &=
    \begin{bmatrix}
        -\omega_{ie}\sin\mu & 0 & -\dfrac{v_E^2}{(R_N + h)^2} \\
        0 & 0 & \dfrac{v_N}{(R_M + h)^2} \\
        -\omega_{ie}\cos\mu - \dfrac{v_E}{(R_N + h)\cos^2\mu} & 0 & \dfrac{v_E\tan\mu}{(R_N + h)^2}
    \end{bmatrix},
    \label{eq:geo_Aap}\\
    \bfF_{\vartheta v} &=
    \begin{bmatrix}
        0 & \dfrac{1}{(R_M + h)} & 0 \\
        -\dfrac{1}{(R_M + h)} & 0 & 0 \\
        0 & -\dfrac{\tan\mu}{(R_N + h)} & 0
    \end{bmatrix},
    \qquad
    \bfF_{\vartheta \vartheta } = -\bfS(\boomega^n_{in}).
    \label{eq:geo_Aav}
\end{align}

The full continuous-time system matrix
$\bfF$ for $\delta\bfx$ in \eqref{eq:geo_error_state} is
\begin{align}
    \bfF =
    \begin{bmatrix}
        \bfF_{pp} & \bfF_{pv} & \bfzero_3 & \bfzero_3 & \bfzero_3 \\
        \bfF_{vp} & \bfF_{vv} & -\bfS(\bfR_{nb}\bff_{ib}^b)
                               & \bfR_{nb} & \bfzero_3 \\
        \bfF_{\vartheta p} & \bfF_{\vartheta v} & \bfF_{\vartheta \vartheta } & \bfzero_3 & -\bfR_{nb} \\
        \bfzero_3 & \bfzero_3 & \bfzero_3 & -p_{\bfa\bfb}\bfI_3 & \bfzero_3 \\
        \bfzero_3 & \bfzero_3 & \bfzero_3 & \bfzero_3 & -p_{\boomega\bfb}\bfI_3
    \end{bmatrix},
    \label{eq:geo_A}
\end{align}
with noise input matrix
\begin{align}
    \bfG =
    \begin{bmatrix}
        \bfzero_3 & \bfzero_3  & \bfzero_3 & \bfzero_3 \\
        \bfR_{nb} & \bfzero_3  & \bfzero_3 & \bfzero_3 \\
        \bfzero_3 & -\bfR_{nb} & \bfzero_3 & \bfzero_3 \\
        \bfzero_3 & \bfzero_3  & \bfI_3    & \bfzero_3 \\
        \bfzero_3 & \bfzero_3  & \bfzero_3 & \bfI_3
    \end{bmatrix},
    \label{eq:geo_G}
\end{align}
where $\bfw = [\boeta_{\bfa}^\top,\,\boeta_{\boomega}^\top,\,
\boeta_{\bfa\bfb}^\top,\,\boeta_{\boomega\bfb}^\top]^\top$.

\subsection{Navigation-Frame Position Updates}

For GNSS position measurement in geodetic coordinates, the measurement equation is simply
\begin{align}
\begin{split}
    \bfy &= \bfh(\bfx) + \bfw
           = \begin{bmatrix}\mu \\ \lambda \\ h\end{bmatrix} + \bfw
    \approx \begin{bmatrix}\hat{\mu} \\ \hat{\lambda} \\ \hat{h}\end{bmatrix}
           + \delta\bfp^g + \bfw
           = \bfh(\hat{\bfx}) + \bfH \delta\bfx + \bfw,
\end{split}
\end{align}
with measurement noise $\bfw \sim \mathcal{N}(\bfzero, \bosigma^2_{\bfw})$.
The corresponding Jacobian is
\begin{align}
    \bfH_\mathrm{pos} = \begin{bmatrix}
        \bfI_3 & \bfzero_3 & \bfzero_3 & \bfzero_3 & \bfzero_3
    \end{bmatrix}.
    \label{eq:geo_H_pos}
\end{align}
As in the additive ECEF formulation in \cref{sec:ECEF_ESKF}, this is a simple identity for the position since both position error $\delta\bfp^g$ and the position measurements are expressed in geodetic coordinates.
This contrasts with the Left-Invariant ECEF formulation in \cref{sec:LI_ECEF} where $\bfH_\mathrm{pos}$
contains $\hat{\bfR}_{eb}$ \eqref{eq:h_pos} in the position sub-block, and the Right-Invariant ECEF formulation in \cref{sec:RI_ECEF}, that includes a non-zero attitude coupling term.

\subsection{Error Injection and Covariance reset}
\label{sec:geo_reset}

\paragraph*{Injection}
In a closed-loop (feedback) implementation, the estimated error-state mean
$\delta\hat{\bfx} = \begin{bmatrix}
    \delta\hat{\bfp}^{g\top} & \delta\hat{\bfv}^{n\top}_{nb} &
    \delta\hat{\bovartheta}_{nb}^{n\top} & \delta\hat{\bfa}_\bfb^{b\top} &
    \delta\hat{\boomega}_\bfb^{b\top}
\end{bmatrix}^\top$
is injected into the nominal state after each update, and the error state is reset to zero.
Position, velocity, and biases are corrected additively, while the tilt
error is applied multiplicatively on $SO(3)$, consistent with the error
definition \eqref{eq:geo_att_err_exact}:
\begin{subequations}
\begin{align}
    \hat{\bfp}^{g}      & \leftarrow \hat{\bfp}^g + \delta\hat{\bfp}^g
        \label{eq:geo_inj_p}\\
    \hat{\bfv}_{nb}^{n} & \leftarrow \hat{\bfv}_{nb}^n + \delta\hat{\bfv}_{nb}^n
        \label{eq:geo_inj_pv}\\
    \hat{\bfq}_{nb}     & \leftarrow \bfq(\delta\hat{\bovartheta}^n_{nb}) \otimes \hat{\bfq}_{nb}
        \quad \text{or} \quad
        \hat{\bfR}_{nb} \leftarrow \Exp(\delta\hat{\bovartheta}^n_{nb})\hat{\bfR}_{nb}
        \label{eq:geo_inj_att}\\
    \hat{\bfa}_\bfb^b     & \leftarrow \hat{\bfa}_\bfb^b + \delta\hat{\bfa}_\bfb^b
        \label{eq:geo_inj_acc_bias}\\
    \hat{\boomega}_\bfb^b & \leftarrow \hat{\boomega}_\bfb^b + \delta\hat{\boomega}_\bfb^b
        \label{eq:geo_inj_gyro_bias}
\end{align}
\label{eq:geo_injection}%
\end{subequations}
The error quaternion is formed from the tilt error in Euler angles,
$\bfq(\delta\hat{\bovartheta}) \approx \big[1,\;\tfrac12\delta\hat{\bovartheta}^\top\big]^\top$,
applied as a \emph{left} (navigation-frame) multiplication in
\eqref{eq:geo_inj_att}.

\paragraph*{ESKF reset}
Because the attitude error is propagated additively as a navigation frame
vector in \eqref{eq:geo_err_kinematics}, the covariance $\bfP$ is the
covariance of $\delta\bfx \in \mathbb{R}^{15}$ and does not need to be transformed from error to nominal state coordinates. 

The reset operation for the error state and covariance is therefore
\begin{align}
    \delta \hbfx &\leftarrow 0 \\
    \hbfP &\leftarrow \boGamma \bfP \boGamma^\top \label{eq:P_reset_geodetic}
\end{align}
with $\boGamma = \bfI$.

%\paragraph{Implementation note.}
%The bias errors $\delta\hat{\bfa}_\bfb$, $\delta\hat{\boomega}_\bfb$ are
%retained in the error state rather than injected into the nominal, so their
%entries in $\delta\hat{\bfx}^{+}$ are carried over unchanged instead of
%being zeroed in \eqref{eq:geo_reset_mean}.

\subsection{Note on Filter Properties}

\paragraph{Fully additive, non-group-affine.}
All three navigation errors are additive in their respective frames,
constituting the classical Pinson error model.
Like the additive ECEF formulation, $\bfF$ depends on the full nominal
state $\bfx = \begin{bmatrix}
        \bfp^{g\top} &
        \bfv_{eb}^{n\top} &
        \delta \bovartheta_{nb}^{n\top} & 
        \bfa_\bfb^{b\top} & \boomega_\bfb^{b\top}
    \end{bmatrix}^\top$ and is thus
trajectory-dependent.

\paragraph{Non-trivial $\bfF$ matrix.}
The geodetic $\bfF$ has non-trivial, trajectory-dependent coupling in all three navigation
subblocks ($\bfF_{pp}$, $\bfF_{vp}$, $\bfF_{\vartheta p}$) because the
transport rate \eqref{eq:geo_rates} depends on both latitude $\mu$
and altitude $h$ through $R_M$, $R_N$, and the trigonometric functions.
The ECEF formulations avoid parts of this complexity by working in a Cartesian frame.

\paragraph{Singularity at the poles.}
The $1/\cos\mu$ factor in $\dot{\lambda}$ and in $\bfF_{pp}$,
$\bfF_{vp}$, $\bfF_{\vartheta p}$ introduces a coordinate singularity at
$\mu = \pm90^\circ$.

\paragraph{Measurement Jacobian}
The simple measurement Jacobian makes navigation-frame measurements well conditioned.
\clearpage
\section{Left-invariant \texorpdfstring{$SE_2(3)$}{SE2(3)} ESKF in ECEF}
\label{sec:LI_ECEF}

This section derives the error state kinematic equations for a left-invariant error formulation (right-multiplication) from \eqref{eq:cont_inertial} as presented in \cite{Hager2025}, and the Jacobian for a simple position update in ECEF coordinates.

\subsection{Error Definition}

The nominal state is composed of the extended Lie Group Pose $\bfT$ \eqref{eq:se23_pose_def} and the \ac{IMU} biases. The corresponding right-multiplicative error $\bozeta = \begin{bmatrix}
    \boxi_1^\top & \boxi_2^\top & \boxi_3^\top
\end{bmatrix}^\top$ is defined as
\begin{align}
    \bfT &:= \hbfT \Exp(\bozeta) \label{eq:T_true_Left}
\end{align}

\begin{align}
\begin{split}
    \Exp(\bozeta) &= \hbfT^{-1} \bfT \\ 
    &= \begin{bmatrix}\begin{array}{c|cc}\hbfR &\hbfv&\hbfp\\\midrule\bfzero_{2\times3} & \multicolumn{2}{c}{\bfI_2} \end{array} \end{bmatrix}^{-1} 
    \begin{bmatrix}\begin{array}{c|cc}\bfR &\bfv&\bfp\\\midrule\bfzero_{2\times3} & \multicolumn{2}{c}{\bfI_2} \end{array} \end{bmatrix}\\ 
    &= \begin{bmatrix}\begin{array}{c|cc}\hbfR^\top & -\hbfR^\top\hbfv&-\hbfR^\top\hbfp\\\midrule\bfzero_{2\times3} & \multicolumn{2}{c}{\bfI_2} \end{array} \end{bmatrix}
    \begin{bmatrix}\begin{array}{c|cc}\bfR &\bfv&\bfp\\\midrule\bfzero_{2\times3} & \multicolumn{2}{c}{\bfI_2} \end{array} \end{bmatrix} \\
    &= \begin{bmatrix}\begin{array}{c|cc}\hbfR^\top \bfR & \hbfR^\top(\bfv - \hbfv) &\hbfR^\top (\bfp - \hbfp) \\\midrule\bfzero_{2\times3} & \multicolumn{2}{c}{\bfI_2} \end{array} \end{bmatrix}- \\ 
\label{eq:error_state_pose}
\end{split}
\end{align}
Thus the attitude, velocity, and position errors are given
\begin{subequations}
\label{eq:err_att_vel_pos}%
\begin{align}
     \log(\hbfR^\top \bfR) &= \boxi_1  \\
     \delta \bfv := \bfv - \hbfv &= \hbfR \bfA(\boxi_1) \boxi_2 \\
     \delta \bfp := \bfp - \hbfp &= \hbfR \bfA(\boxi_1) \boxi_3.
\end{align} %
\end{subequations}%
Taking a first-order approximation of \cref{eq:Exp_SE23} and inserting for \cref{eq:error_state_pose} yields the extended pose error

\begin{subequations}
\begin{alignat}{3}
    \boxi_1 &\approx (\hbfR^\top \bfR- \bfI_3) &&\Leftrightarrow \bfR &&\approx \hbfR\left( \bfI_3 + \bfS( \boxi_1)\right)  \label{eq:att_err_1st_ord}  \\
    \boxi_2 &\approx \hbfR^\top(\bfv - \hbfv) &&\Leftrightarrow \bfv && \approx\hbfv  + \hbfR \boxi_2 \label{eq:vel_err_1st_ord} 
   \\
    \boxi_3 &\approx \hbfR^\top (\bfp - \hbfp) &&\Leftrightarrow \bfp && \approx\hbfp  + \hbfR \boxi_3 \label{eq:pos_err_1st_ord} 
\end{alignat}
\label{eq:ext_pos_err}%
\end{subequations}%c026
%where 
%The extended pose error is given by $\boxi = \begin{bmatrix}\boxi_1 & \boxi_2 & \boxi_3 \end{bmatrix}^\top$ where $\delta \bfR = \exp(\bfS(\boxi_1)) $ is the attitude error, $ \delta \bfv = \hbfR \bfA(\boxi_1) \boxi_2$ the velocity error, and $ \delta \bfp = \hbfR \bfA(\boxi_1) \boxi_3$ the position error. 
%The
\begin{equation}
   \delta \bfx = \begin{bmatrix}
\boxi_1^\top & \boxi_2^\top & \boxi_3^\top & \delta \bfa_\bfb^{b \top} & \delta \boomega_\bfb^{b \top} 
\end{bmatrix}^\top,
\label{eq:error_state_total_left}
\end{equation}
where $\boxi_1, \boxi_2, \boxi_3 \in \mathbb{R}^3$ are the local body-frame attitude, velocity, and position errors, respectively.

\subsection{Error State Kinematics}
\label{app:err_kinematics_left_invariant}

\subsubsection{Continuous-time error state kinematics}

\begin{subequations}
\begin{align}
\begin{split}
\dot{\boxi}_1 &= -\bfS( \hboomega^b_{eb} ) \boxi_1 - \delta \boomega_{\bfb}^b - \boeta^b_{\boomega \bfb}
\end{split}\\
\begin{split}
\dot{\boxi}_2 &= - ( \bfS( \hboomega^b_{eb} ) + 2 \bfS(\boomega^b_{ie}) ) \boxi_2  - \bfS(\hbff^b_{ib} ) \boxi_1 -  \delta\bfa_{\bfb}^b - \boeta^b_{\bfa \bfb} 
\end{split}\\
\begin{split}
\dot{\boxi}_3 &= -\bfS( \hboomega^b_{eb} ) \boxi_3 + \boxi_2
\end{split}\\
\begin{split}
\delta \dot{\bfa}_{\bfb}^b &= - p_{\bfa \bfb} \delta\bfa_{\bfb}^b + \boeta^b_{\bfa \bfb}
\end{split}\\
\begin{split}
\delta \dot{\boomega}_{\bfb}^b &= - p_{\boomega \bfb} \delta\boomega_{\bfb }^b + \boeta^b_{\boomega \bfb}. 
\end{split}
\end{align}
\label{eq:err_kinematics}%
\end{subequations}

\subsubsection{System Matrix}
\label{ssec:F_LI}

The full continuous-time system matrix 
$\bfF$ for $\delta\bfx = \begin{bmatrix}\boxi_1^\top & \boxi_2^\top & \boxi_3^{b \top} & \delta\bfa_\bfb^{b \top} & \delta\boomega_\bfb^\top\end{bmatrix}^\top$ is:

With bias coupling:
\begin{align}
\label{eq:F_LI}
    \bfF = 
    \begin{bmatrix}
        -\bfS(\hboomega^b_{eb}) & \bfzero_3 & \bfzero_3 & \bfzero_3 & -\bfI_3 \\[4pt]
        -\bfS(\hbff^b_{ib})     & -(\bfS(\hboomega^b_{eb}) + 2\bfS(\boomega^b_{ie})) & \bfzero_3 & -\bfI_3 & \bfzero_3 \\[4pt]
        \bfzero_3               & \bfI_3    & -\bfS(\hboomega^b_{eb}) & \bfzero_3 & \bfzero_3 \\[4pt]
        \bfzero_3 & \bfzero_3 & \bfzero_3 & -p_{\bfa\bfb}\bfI_3 & \bfzero_3 \\[4pt]
        \bfzero_3 & \bfzero_3 & \bfzero_3 & \bfzero_3 & -p_{\boomega\bfb}\bfI_3
    \end{bmatrix}
\end{align}

With PVA covariance propagation decoupled from bias propagation:
\begin{align}
    \bfF = 
    \begin{bmatrix}
        -\bfS(\hboomega^b_{eb}) & \bfzero_3 & \bfzero_3 & \bfzero_3 & \bfzero_3 \\[4pt]
        -\bfS(\hbff^b_{ib})     & -(\bfS(\hboomega^b_{eb}) + 2\bfS(\boomega^b_{ie})) & \bfzero_3 & \bfzero_3 & \bfzero_3 \\[4pt]
        \bfzero_3               & \bfI_3    & -\bfS(\hboomega^b_{eb}) & \bfzero_3 & \bfzero_3 \\[4pt]
        \bfzero_3 & \bfzero_3 & \bfzero_3 & -p_{\bfa\bfb}\bfI_3 & \bfzero_3 \\[4pt]
        \bfzero_3 & \bfzero_3 & \bfzero_3 & \bfzero_3 & -p_{\boomega\bfb}\bfI_3
    \end{bmatrix}
\end{align}

Noise coupling matrix:
\begin{align}
    \bfG = 
    \begin{bmatrix}
        \bfzero_3 & -\bfI_3 & \bfzero_3 & \bfzero_3 \\[4pt]
        -\bfI_3 & \bfzero_3 & \bfzero_3 & \bfzero_3 \\[4pt]
        \bfzero_3 & \bfzero_3 & \bfzero_3 & \bfzero_3 \\[4pt]
        \bfzero_3 & \bfzero_3 & \bfI_3 & \bfzero_3 \\[4pt]
        \bfzero_3 & \bfzero_3 & \bfzero_3 & \bfI_3
    \end{bmatrix}
\end{align}

\subsubsection{Derivation of the Error State Kinematics}
The linearized dynamics of the error state in eq. \ref{eq:err_kinematics} are derived from \cref{eq:ext_pos_err} analogous to Chapter 5.3 in Sola's ESKF formulation \cite{Sola2021}.
The time-derivative of $\hbfR_{eb}$ is 
\begin{align}
\begin{split}
    \dot{\hbfR}_{eb} &= \hbfR_{eb} \bfS( \hboomega^b_{ib} ) - \bfS( \boomega^e_{ie} ) \hbfR_{eb} \\
    &= \hbfR_{eb} \bfS( \hboomega^b_{ib} ) - \hbfR_{eb} \hbfR_{eb}^\top \bfS( \boomega^e_{ie} ) \hbfR_{eb} \\
    %
    %&= \hbfR_{eb} \bfS( \hboomega^b_{ib} ) - \hbfR_{eb} \bfS( \hboomega^b_{ie} ) %\\
    %
    &= \hbfR_{eb} ( \bfS( \hboomega^b_{ib} ) - \bfS( \hboomega^b_{ie} ) ) = \hbfR_{eb} \bfS( \hboomega^b_{eb} )
\end{split}
\label{eq:hbfR_dot_left}
\end{align}
and will be useful in the following.
\paragraph{Attitude Error}
From \cite{sola_quat} we know that
\begin{align}
%\begin{split}
\dot{\boxi}_1 &= -\bfS( \hboomega^b_{eb} ) \boxi_1 - \delta \boomega_{\bfb}^b - \boeta^b_{\boomega},
%\end{split}\\
\end{align}
where $\hboomega^b_{eb} = \hboomega^b_{ib} - \hbfR_{eb}^\top\boomega^e_{ie}$ is the body rate relative to the ECEF frame. 
\paragraph{Velocity Error}
\label{p:vel_err_RI}
From the definition of the linearized error state from \cref{eq:vel_err_1st_ord}, we know that
%
%\begin{align}
%    \bfv^e_{eb} &\approx \hbfv^e_{eb} + \hbfR_{eb}\boxi_2,
%\end{align}
%
%which is resulting in
%
\begin{align*}
    \boxi_2 &\approx  \hbfR_{eb}^\top (\bfv^e_{eb}- \hbfv^e_{eb})
\end{align*}
for the velocity error. Its time derivative is then given by
\begin{align}
\begin{split}
\dboxi_2 &\approx \dot{\hbfR}_{eb}^\top (\bfv^e_{eb}- \hbfv^e_{eb}) + \hbfR_{eb}^\top (\dot{\bfv}^e_{eb}- \dot{\hbfv}^e_{eb})  \\
&= - \bfS( \hboomega^b_{eb} ) \hbfR_{eb}^\top (\bfv^e_{eb}- \hbfv^e_{eb})+ \hbfR_{eb}^\top ( - \bfS(2\boomega^e_{ie}) \bfv^e_{eb} %\notag \\ &
+ \bfR_{eb}(\bff^b_m-\hbfa^b_\bfb - \boeta^b_{\bfa} -\boeta^b_{\bfa \bfb})  \\ &\quad + \bfg^e  - \bfS^2(\boomega^e_{ie}) \bfp^e_{eb} %\notag  \\
%& 
+ \bfS(2\boomega^e_{ie}) \bfv^e_{eb} - \hbfR_{eb}(\bff^b_m-\hbfa^b_\bfb) - \bfg^e  + \bfS^2(\hboomega^e_{ie}) \hbfp^e_{eb})  \\
&\approx - \bfS( \hboomega^b_{eb} ) \boxi_2 - 2 \hbfR_{eb}^\top \bfS(\boomega^e_{ie}) (\bfv^e_{eb} - \hbfv^e_{eb})   \\ 
& \quad + \hbfR_{eb}^\top \bfR_{eb} (\bfI + \bfS(\boxi_1))(\bff^b_m-\hbfa^b_\bfb - \delta \bfa_\bfb^b-\boeta^b_{\bfa}) %\notag  \\ 
%
%&
- \hbfR_{eb}^\top \hbfR_{eb} (\bff^b_m-\hbfa^b_\bfb)  \\
&\approx - \bfS( \hboomega^b_{eb} ) \boxi_2 - 2 \hbfR_{eb}^\top \bfS(\boomega^e_{ie})\hbfR_{eb}\hbfR_{eb}^\top (\bfv^e_{eb} - \hbfv^e_{eb})   \\
& \quad + \bfS(\boxi_1) (\bff^b_m-\hbfa^b_\bfb - \delta \bfa_\bfb^b-\boeta^b_{\bfa}) - \delta \bfa_\bfb^b - \boeta^b_{\bfa}  \\
&\approx - \bfS( \hboomega^b_{eb} ) \boxi_2 - 2 \bfS(\boomega^b_{ie}) \boxi_2 - \bfS(\hbff^b_{ib}) \boxi_1 - \delta \bfa_\bfb^b - \boeta^b_{\bfa},
\end{split}
\end{align}
where we have used the inertial equation for the velocity in the first step, $\bfS^2(\boomega^e_{ie})\bfp^e_{eb} \approx \bfS^2(\hboomega^e_{ie})\hbfp^e_{eb} $ in the second step, $\hbfR_{eb} \bfR_{eb} \approx \bfI_3$ in the third step, and finally $\bfS(\boxi_1)\delta \bfa_\bfb^b \approx 0 $, $\bfS(\boxi_1)\boeta^b_{\bfa \bfb} \approx 0 $ and $\hbff^b_{ib} = \bff^b_m-\hbfa^b_\bfb$. 

\paragraph{Position Error}
From \cref{eq:pos_err_1st_ord}, the left-invariant position error satisfies 
$\boxi_3 \approx \hbfR_{eb}^\top(\bfp^e_{eb} - \hbfp^e_{eb})$.
Differentiating with respect to time:
\begin{align}
    \dot{\boxi}_3 
    &= \dot{\hbfR}_{eb}^\top(\bfp^e_{eb} - \hbfp^e_{eb}) 
     + \hbfR_{eb}^\top(\bfv^e_{eb} - \hbfv^e_{eb}).
\end{align}
Using $\dot{\hbfR}_{eb}^\top = -\bfS(\hboomega^b_{eb})\hbfR_{eb}^\top$ \cref{eq:hbfR_dot_left}
and $\hbfR_{eb}^\top(\bfv^e_{eb} - \hbfv^e_{eb}) = \boxi_2$ \cref{eq:vel_err_1st_ord}:
\begin{align}
\begin{split}
    \dot{\boxi}_3 
    &= -\bfS(\hboomega^b_{eb})\underbrace{\hbfR_{eb}^\top(\bfp^e_{eb} - \hbfp^e_{eb})}_{\boxi_3} 
       + \boxi_2 
       + \bfS(\boxi_1)\hbfR_{eb}^\top\hbfv^e_{eb}. \\
    &= -\bfS(\hboomega^b_{eb})\boxi_3 + \boxi_2.
\end{split}    
\end{align}
The position error kinematics contain no noise or bias terms at first order, unlike the right-invariant formulation, 
Because of the position error being in body frame, no gyroscope coupling appears here.

\subsection{Navigation-Frame Position Updates}

Assuming position measurements given in the filter frame, ECEF

\begin{align}
\begin{split}
    \bfy &= \bfh(\bfx) + \bfw = \bfp_{eb}^e + \bfw \\
    &\approx \hbfp_{eb}^e + \hbfR_{eb} \boxi_3 + \bfw \\ 
    &= \bfh(\hbfx) + \bfH \delta \bfx + \bfw,
    \end{split} \label{eq:h_pos}
\end{align}

with measurement noise $\bfw \sim \mathcal{N} (\bm 0, \bosigma_{\bfw}^2)$. Here \cref{eq:ext_pos_err} was used to express $\bfy$ with respect to the errors and the nominal state. From \cref{eq:h_pos}, the measurement Jacobian $\bfH \equiv \left. \frac{\partial \bfh}{\partial \delta \bfx} \right|_{\delta \bfx = \bm 0} $ can be read off directly:
 \begin{align}
     \bfH_\text{pos} = \begin{bmatrix}
        \bfzero_{1 \times 3} & \bfzero_{1 \times 3} & \hbfR_{eb} & \bfzero_{1 \times 3} & \bfzero_{1 \times 3}
    \end{bmatrix}.
 \end{align}

The rotation matrix in the position block maps the world-frame measurement to the body frame of the error state.
\subsection{Error Injection and Covariance reset}

\paragraph*{Injection}
The injection of the error into the total state,
\begin{align}
    \bfx \leftarrow \hbfx \oplus \delta \hbfx,
\end{align}
follows directly from \eqref{eq:ext_pos_err} and is then given by the compositions 
\begin{subequations}
\begin{align}
    \hbfR_{eb} &\leftarrow \hbfR_{eb} \Exp (\hboxi_1) \label{eq:att_inj}\\
    \hbfv^e_{eb} &\leftarrow  \hbfv^e_{eb} + \hbfR_{eb} \bfA(\hboxi_1) \hboxi_2 \\
    \hbfp^e_{eb} &\leftarrow  \hbfp^e_{eb} + \hbfR_{eb} \bfA(\hboxi_1) \hboxi_3 \label{eq:pos_inj}\\
    \hbfa_\bfb^b &\leftarrow  \hbfa_\bfb^b + \delta \hbfa_{\bfb}^b \\
    \hboomega^b_\bfb &\leftarrow  \hboomega^b_\bfb + \delta \hboomega_{\bfb}^b,
\end{align}
\end{subequations}
where \crefrange{eq:att_inj}{eq:pos_inj} are given by \cref{eq:err_att_vel_pos}, and the bias injection is simply an addition.
\paragraph*{ESKF reset}
\label{ssec:eskf_reset_li}
The reset operation for the error state and covariance is
\begin{align}
    \delta \hbfx &\leftarrow 0 \\
    \hbfP &\leftarrow \boGamma \bfP \boGamma^\top \label{eq:P_reset_li}
\end{align}
similar to \cite{sola_quat} where in (\ref{eq:P_reset_li}), $\boGamma$ is the Jacobian of the error reset function \eqref{eq:reset_fct} as given by \eqref{eq:cov_reset}

\begin{align}
    \boGamma = \left. \frac{\partial \gamma}{\partial \delta \bfx}\right|_{\delta \hbfx} = \begin{bmatrix}
       \bfJ_r^{SE_2(3)}(\bozeta) & \bfzero_{9\times6}\\
        \bfzero_{6\times9} & \bfI_6  \\
    \end{bmatrix},
\end{align}
where the full expression for $\bfJ_r^{SE_2(3)}$ or a first-order approximation of the $SO(3)$ Jacobian for the attitude block only,
\begin{align}
\bfJ_r^{SO(3)}(\boxi_1) \approx \bfI_3 - \frac{1}{2} \bfS(\hboxi_1),
 \end{align}

can be used.

\subsection{Note on Filter Properties}

\paragraph{Group-affine error propagation.}
$\bfF$ in \cref{eq:F_LI} does not contain any dependency on the state ($\hbfp, \hbfv, \hat{\bovartheta}$ and is thus trajectory-independent in contrast to all other filters presented here.
Together with a full-order covariance reset, the covariance estimate of this formulation is near-exact.

\paragraph{Numerical conditioning.}
The largest coupling block is
$\bfF_{v \vartheta} = \|\bfS(\bff_{ib}^b)\| \sim g_0 \approx 9.8$~m/s$^2$, similar to the geodetic/local-navigation frame filter.

\paragraph{Measurement Jacobian}
The additional rotation matrix in the position coupling block of the Measurement Jacobian introduces a coupling of the world-frame measurements with the current attitude estimate. This can destabilize the filter when attitude estimates are insufficiently accurate.
\clearpage
\section{Right-invariant \texorpdfstring{$SE_2(3)$}{SE2(3)} ESKF in ECEF}
\label{sec:RI_ECEF}

This section derives the error state kinematic equations for a right-invariant error formulation (left-multiplication) from \eqref{eq:cont_inertial} analogous to \cite{Hager2025}, and the Jacobian for a simple position update in ECEF coordinates.

\subsection{Error Definition}

The nominal state is composed of the extended Lie Group Pose $\bfT$ \eqref{eq:se23_pose_def} and the \ac{IMU} biases. The corresponding left-multiplicative error $\bozeta = \begin{bmatrix}
    \boxi_1^\top & \boxi_2^\top & \boxi_3^\top
\end{bmatrix}^\top$ is defined as

\begin{align}
    \bfT &:= \Exp(\bozeta) \hbfT \label{eq:T_true_Right}
\end{align}

\begin{align}
    \begin{split} 
    \Exp(\bozeta) &= \bfT \hbfT^{-1} \\ 
    &= \begin{bmatrix}\begin{array}{c|cc}\bfR &\bfv&\bfp\\\midrule\bfzero_{2\times3} & \multicolumn{2}{c}{\bfI_2} \end{array} \end{bmatrix}
    \begin{bmatrix}\begin{array}{c|cc}\hbfR^\top & -\hbfR^\top\hbfv&-\hbfR^\top\hbfp\\\midrule\bfzero_{2\times3} & \multicolumn{2}{c}{\bfI_2} \end{array} \end{bmatrix} \\ 
    &= \begin{bmatrix}\begin{array}{c|cc}\bfR\hbfR^\top & \bfv - \bfR\hbfR^\top\hbfv &\bfp - \bfR\hbfR^\top\hbfp \\\midrule\bfzero_{2\times3} & \multicolumn{2}{c}{\bfI_2} \end{array} \end{bmatrix} 
    \end{split}, 
    \label{eq:error_state_pose_right}
\end{align}
leading to the definition of attitude velocity and position errors 

\begin{subequations}
\begin{align}
     \bfS(\boxi_1) &= \log(\bfR\hbfR^\top)  \label{eq:att_err_right}\\
     \boxi_2 &:= \bfv - \bfR\hbfR^\top\hbfv \label{eq:vel_err_right} \\
     \boxi_3 &:= \bfp - \bfR\hbfR^\top\hbfp. \label{eq:pos_err_right}
\end{align} \label{eq:err_att_vel_pos_right}%
\end{subequations}%

Taking a first-order approximation ($\bfR\hbfR^\top \approx \bfI_3 + \bfS(\boxi_1)$) and inserting for \cref{eq:error_state_pose_right} yields the extended pose error approximations:
\begin{subequations}
\begin{alignat}{3}
    \bfS(\boxi_1) &\approx (\bfR\hbfR^\top - \bfI_3) &&\Leftrightarrow \bfR &&\approx \left( \bfI_3 + \bfS( \boxi_1)\right) \hbfR  \label{eq:att_err_1st_ord_right}  \\
    \boxi_2 &\approx \bfv - \hbfv - \bfS(\boxi_1)\hbfv &&\Leftrightarrow \bfv && \approx \hbfv + \boxi_2 + \bfS(\boxi_1)\hbfv \label{eq:vel_err_1st_ord_right}  
   \\
    \boxi_3 &\approx \bfp - \hbfp - \bfS(\boxi_1)\hbfp &&\Leftrightarrow \bfp && \approx \hbfp + \boxi_3 + \bfS(\boxi_1)\hbfp \label{eq:pos_err_1st_ord_right}  
\end{alignat}
\label{eq:ext_pos_err_right}%
\end{subequations}%

The error-state vector including biases has the same form as for the Left-Invariant filter cf. \cref{eq:error_state_total_left}:
\begin{equation}
   \delta \bfx = \begin{bmatrix}
\boxi_1^\top & \boxi_2^\top & \boxi_3^\top & \delta \bfa_\bfb^\top & \delta \boomega_\bfb^\top 
\end{bmatrix}^\top,
\label{eq:error_state_total_right}
\end{equation}
now with $\boxi_1, \boxi_2, \boxi_3 \in \mathbb{R}^3$ defined as global ECEF-frame attitude, velocity, and position errors, respectively.

\subsection{Error State Kinematics}
\label{app:err_kinematics_right}

\subsubsection{Continous-time error state kinematics}

\begin{subequations}
\begin{align}
\dot{\boxi}_1 &= -\bfS(\boomega^e_{ie})\boxi_1 
                 - \hbfR_{eb}\delta\boomega_\bfb 
                 - \hbfR_{eb}\boeta^b_{\boomega} \\
\dot{\boxi}_2 &= -2\bfS(\boomega^e_{ie})\boxi_2 
                 + \bfS(\bfg^e)\boxi_1 
                 - \hbfR_{eb}\delta\bfa_\bfb 
                 - \hbfR_{eb}\boeta^b_{\bfa}
                 - \bfS(\hbfv^e_{eb})\hbfR_{eb}\delta\boomega_\bfb
                 - \bfS(\hbfv^e_{eb})\hbfR_{eb}\boeta^b_{\boomega} \\
\dot{\boxi}_3 &= \boxi_2 
                 - \bfS(\hbfp^e_{eb})\bfS(\boomega^e_{ie})\boxi_1
                 - \bfS(\hbfp^e_{eb})\hbfR_{eb}\delta\boomega_\bfb
                 - \bfS(\hbfp^e_{eb})\hbfR_{eb}\boeta^b_{\boomega} \\
\delta \dot{\bfa}_{\bfb}^b &= - p_{\bfa \bfb} \delta\bfa_{\bfb}^b + \boeta^b_{\bfa \bfb} \\
\delta \dot{\boomega}_{\bfb}^b &= - p_{\boomega \bfb} \delta\boomega_{\bfb}^b + \boeta^b_{\boomega \bfb}. 
\end{align}
\label{eq:err_kinematics_right}%
\end{subequations}

\subsubsection{System matrix}

With bias coupling terms
\begin{align}
    \bfF = 
    \begin{bmatrix}
        -\bfS(\boomega^e_{ie}) & \bfzero_3 & \bfzero_3 & \bfzero_3 & -\hbfR_{eb} \\[4pt]
        \bfS(\bfg^e)           & -2\bfS(\boomega^e_{ie}) & \bfzero_3 & -\hbfR_{eb} & -\bfS(\hbfv^e_{eb})\hbfR_{eb} \\[4pt]
        -\bfS(\hbfp^e_{eb})\bfS(\boomega^e_{ie}) & \bfI_3 & \bfzero_3 & \bfzero_3 & -\bfS(\hbfp^e_{eb})\hbfR_{eb} \\[4pt]
        \bfzero_3 & \bfzero_3 & \bfzero_3 & -p_{\bfa\bfb}\bfI_3 & \bfzero_3 \\[4pt]
        \bfzero_3 & \bfzero_3 & \bfzero_3 & \bfzero_3 & -p_{\boomega\bfb}\bfI_3
    \end{bmatrix}
    \label{eq:A_full_RI}
\end{align}

With PVA covariance propagation decoupled from bias propagation:

\begin{align}
    \bfF = 
    \begin{bmatrix}
        -\bfS(\boomega^e_{ie}) & \bfzero_3 & \bfzero_3 & \bfzero_3 & \bfzero_3 \\[4pt]
        \bfS(\bfg^e)           & -2\bfS(\boomega^e_{ie}) & \bfzero_3 & \bfzero_3 & \bfzero_3 \\[4pt]
        -\bfS(\hbfp^e_{eb})\bfS(\boomega^e_{ie}) & \bfI_3 & \bfzero_3 & \bfzero_3 & \bfzero_3 \\[4pt]
        \bfzero_3 & \bfzero_3 & \bfzero_3 & -p_{\bfa\bfb}\bfI_3 & \bfzero_3 \\[4pt]
        \bfzero_3 & \bfzero_3 & \bfzero_3 & \bfzero_3 & -p_{\boomega\bfb}\bfI_3
    \end{bmatrix}
\end{align}

Noise coupling matrix: 

With full noise coupling
\begin{align}
    \bfG = 
    \begin{bmatrix}
        \bfzero_3 & -\hbfR_{eb} & \bfzero_3 & \bfzero_3 \\[4pt]
        -\hbfR_{eb} & -\bfS(\hbfv^e_{eb})\hbfR_{eb} & \bfzero_3 & \bfzero_3 \\[4pt]
        \bfzero_3 & -\bfS(\hbfp^e_{eb})\hbfR_{eb} & \bfzero_3 & \bfzero_3 \\[4pt]
        \bfzero_3 & \bfzero_3 & \bfI_3 & \bfzero_3 \\[4pt]
        \bfzero_3 & \bfzero_3 & \bfzero_3 & \bfI_3
    \end{bmatrix}
    \label{eq:G_full_RI}
\end{align}

Truncated (log-linear)
\begin{align}
    \bfG = 
    \begin{bmatrix}
        \bfzero_3 & -\hbfR_{eb} & \bfzero_3 & \bfzero_3 \\[4pt]
        -\hbfR_{eb} & \bfzero_3 & \bfzero_3 & \bfzero_3 \\[4pt]
        \bfzero_3 & \bfzero_3 & \bfzero_3 & \bfzero_3 \\[4pt]
        \bfzero_3 & \bfzero_3 & \bfI_3 & \bfzero_3 \\[4pt]
        \bfzero_3 & \bfzero_3 & \bfzero_3 & \bfI_3
    \end{bmatrix}
\end{align}

The gyroscope noise has additional couplings into velocity and position errors that violate the independency of the error dynamics on the current total state. This is why this formulation including gyroscope and accelerometer biases consitutes an \textit{imperfect} invariant filter \cite{barrau_thesis}. 
They need to be truncated to preserve the group structure.

\subsubsection{Derivation of the Error State Kinematics}

The linearized dynamics of the right-invariant error state in \cref{eq:err_kinematics_right} are derived from the left-multiplied error definitions in \cref{eq:ext_pos_err_right}. In this framework, the extended pose error resides in the global navigation frame (ECEF).
The time-derivative of the nominal attitude $\hbfR_{eb}$ is the standard propagation:
\begin{align}
    \dot{\hbfR}_{eb} &= -\bfS(\boomega^e_{ie})\hbfR_{eb} + \hbfR_{eb}\bfS(\hboomega^b_{ib})
\end{align}
where $\hboomega^b_{ib} = \boomega^b_m - \hboomega^b_\bfb$.

\paragraph{Attitude Error}
The true attitude relates to the right-invariant error via $\bfR_{eb} = \exp(\bfS(\boxi_1))\hbfR_{eb}$. Taking the time derivative of both sides yields:

\begin{align}
    \begin{split}
        \dot{\bfR}_{eb}  &= -\bfS(\boomega^e_{ie})\bfR_{eb} + \bfR_{eb}\bfS(\boomega^b_{ib}) \\
        &= -\bfS(\boomega^e_{ie})(\bfI_3 + \bfS(\boxi_1))\hbfR_{eb} + (\bfI_3 + \bfS(\boxi_1))\hbfR_{eb}\bfS(\hboomega^b_{ib} - \delta \boomega_\bfb^b - \delta \boeta_{\boomega}^b  ) \\
        &= -\bfS(\boomega^e_{ie})\hbfR_{eb} -\bfS(\boomega^e_{ie}) \bfS(\boxi_1)\hbfR_{eb} + \hbfR_{eb}\bfS(\hboomega^b_{ib} - \delta \boomega_\bfb^b - \delta \boeta_{\boomega}^b) + \bfS(\boxi_1)\hbfR_{eb}\bfS(\hboomega^b_{ib} - \delta \boomega_\bfb^b - \delta \boeta_{\boomega}^b  ) %\\
        %
        %&\approx \bfS(\dot{\boxi}_1)\hbfR_{eb} + (\bfI_3 + \bfS(\boxi_1))\left(-\bfS(\boomega^e_{ie})\hbfR_{eb} + \hbfR_{eb}\bfS(\hboomega^b_{ib})\right) \\
        %
        %&=-\bfS(\boomega^e_{ie})\hbfR_{eb} + \hbfR_{eb}\bfS(\hboomega^b_{ib}) + \bfS(\dot{\boxi}_1)\hbfR_{eb}  -\bfS(\boxi_1)\bfS(\boomega^e_{ie})\hbfR_{eb} + \bfS(\boxi_1)\hbfR_{eb}\bfS(\hboomega^b_{ib})   \\
        \label{eq:R_eb_dot_right_invariant_temp}
    \end{split}
\end{align}
Approximating 
\begin{align}
    \bfR_{eb} &= \exp(\bfS(\boxi_1))\hbfR_{eb} \\
    &\approx (\bfI_3 + \bfS(\boxi_1))\hbfR_{eb} \\
    \Rightarrow \dot{\bfR}_{eb} &\approx \overbrace{(\bfI_3 + \bfS(\boxi_1))}^{\dot{}}\hbfR_{eb} + (\bfI_3 + \bfS(\boxi_1))\dot{\hbfR}_{eb}  \\
    &= \bfS(\dot{\boxi}_1)\hbfR_{eb} + \dot{\hbfR}_{eb} + \bfS(\boxi_1)\dot{\hbfR}_{eb} \label{eq:R_eb_dot_right_invariant_temp2}
\end{align}
Putting \cref{eq:R_eb_dot_right_invariant_temp2} on the left side of  \cref{eq:R_eb_dot_right_invariant_temp} results in
\begin{align}
    \begin{split}
            \bfS(\dot{\boxi}_1)\hbfR_{eb} + \dot{\hbfR}_{eb} + \bfS(\boxi_1)\dot{\hbfR}_{eb} &= 
            -\bfS(\boomega^e_{ie})\hbfR_{eb} 
            -\bfS(\boomega^e_{ie}) \bfS(\boxi_1)\hbfR_{eb} +
            \hbfR_{eb}\bfS(\hboomega^b_{ib} - \delta \boomega_\bfb^b - \delta \boeta_{\boomega}^b) \\
            & \quad + \bfS(\boxi_1)\hbfR_{eb}\bfS(\hboomega^b_{ib} - \delta \boomega_\bfb^b - \delta \boeta_{\boomega}^b  ) %\\
    \end{split}
\end{align}
This can further be simplified by removing 
\begin{align*}
    \dot{\hbfR}_{eb} &= -\bfS(\boomega^e_{ie})\hbfR_{eb}  + \hbfR_{eb}\bfS(\hboomega^b_{ib}) 
\end{align*}
from the the equation since this is only associated with the propagation of the attitude. Then one is left with
\begin{align}
    \begin{split}
            \bfS(\dot{\boxi}_1)\hbfR_{eb} + \bfS(\boxi_1)\dot{\hbfR}_{eb} &= 
            -\bfS(\boomega^e_{ie}) \bfS(\boxi_1)\hbfR_{eb} 
            -\hbfR_{eb}\bfS( \delta \boomega_\bfb^b + \delta \boeta_{\boomega}^b) \\
            & \quad + \bfS(\boxi_1)\hbfR_{eb}\bfS(\hboomega^b_{ib} - \delta \boomega_\bfb^b - \delta \boeta_{\boomega}^b  ) \\
            \Rightarrow \bfS(\dot{\boxi}_1)\hbfR_{eb} &= - \bfS(\boxi_1)\dot{\hbfR}_{eb} -\bfS(\boomega^e_{ie}) \bfS(\boxi_1)\hbfR_{eb} 
            -\hbfR_{eb}\bfS( \delta \boomega_\bfb^b + \delta \boeta_{\boomega}^b) \\
            & \quad + \bfS(\boxi_1)\hbfR_{eb}\bfS(\hboomega^b_{ib} - \delta \boomega_\bfb^b - \delta \boeta_{\boomega}^b  ) \\
            &= -\bfS(\boxi_1)\hbfR_{eb}\bfS(\hboomega^b_{ib}) -\bfS(\boomega^e_{ie}) \bfS(\boxi_1)\hbfR_{eb} 
            -\hbfR_{eb}\bfS( \delta \boomega_\bfb^b + \delta \boeta_{\boomega}^b) \\
            & \quad + \bfS(\boxi_1)\hbfR_{eb}\bfS(\hboomega^b_{ib}) - \bfS(\boxi_1)\hbfR_{eb}\bfS(\delta \boomega_\bfb^b + \delta \boeta_{\boomega}^b  ) \\
            &= -\bfS(\boomega^e_{ie}) \bfS(\boxi_1)\hbfR_{eb} 
            -\hbfR_{eb}\bfS( \delta \boomega_\bfb^b + \delta \boeta_{\boomega}^b) \\
            & \quad - \bfS(\boxi_1)\hbfR_{eb}\bfS(\delta \boomega_\bfb^b + \delta \boeta_{\boomega}^b  ) \\
            &\approx
            -\bfS(\boomega^e_{ie}) \bfS(\boxi_1)\hbfR_{eb} 
            -\hbfR_{eb}\bfS( \delta \boomega_\bfb^b + \delta \boeta_{\boomega}^b)  \\
            \Rightarrow \bfS(\dot{\boxi}_1) &\approx  -\bfS(\boomega^e_{ie}) \bfS(\boxi_1)\hbfR_{eb}\hbfR_{eb}^\top
            -\hbfR_{eb}\bfS( \delta \boomega_\bfb^b + \delta \boeta_{\boomega}^b)\hbfR_{eb}^\top  \\
            &= -\bfS(\boomega^e_{ie}) \bfS(\boxi_1)
            -\bfS( \hbfR_{eb} \delta \boomega_\bfb^b + \hbfR_{eb} \delta \boeta_{\boomega}^b) 
    \end{split}
\end{align}
Extracting the vector components directly yields the orientation error kinematics:
\begin{align}
    \dot{\boxi}_1 &= 
    -\bfS(\boomega^e_{ie})\boxi_1 
    - \hbfR_{eb}\delta \boomega_\bfb^b
    - \hbfR_{eb}\boeta_\omega,
    \label{eq:dxi_1_ri}
\end{align}

simplifying to 
\begin{align}
    \dot{\boxi}_1 &= 
    -\bfS(\boomega^e_{ie})\boxi_1 
\end{align}
without the noise terms.

\paragraph{Velocity Error}
From \cref{eq:vel_err_right}, the right-invariant velocity error is defined directly 
in the ECEF frame as $\boxi_2 = \bfv^e_{eb} - \bfR_{eb}\hbfR_{eb}^\top\hbfv^e_{eb}$. 
To first order, this simplifies to 
\begin{align}
    \boxi_2 \approx \bfv^e_{eb} - \hbfv^e_{eb} - \bfS(\boxi_1)\hbfv^e_{eb}.
\end{align}
Differentiating with respect to time yields:
\begin{align}
    \dot{\boxi}_2 \approx 
    \dot{\bfv}^e_{eb} 
    - \dot{\hbfv}^e_{eb} 
    - \bfS(\dot{\boxi}_1)\hbfv^e_{eb} 
    - \bfS(\boxi_1)\dot{\hbfv}^e_{eb}.
\end{align}
Substituting the true and estimated continuous velocity kinematics from \cref{eq:dv}:
\begin{align}
\begin{split}
    \dot{\boxi}_2 
    &\approx \left(-2\bfS(\boomega^e_{ie})\bfv^e_{eb} + \bfR_{eb}\bff^b_{ib} + \bfg^e(\bfp^e)\right) \\
    &\quad - \left(-2\bfS(\boomega^e_{ie})\hbfv^e_{eb} + \hbfR_{eb}\hbff^b_{ib} + \bfg^e(\hbfp^e)\right) \\
    &\quad - \bfS(\dot{\boxi}_1)\hbfv^e_{eb} 
           - \bfS(\boxi_1)\left(-2\bfS(\boomega^e_{ie})\hbfv^e_{eb} + \hbfR_{eb}\hbff^b_{ib} + \bfg^e(\hbfp^e)\right).
\end{split}
\end{align}
Next, the true velocity is replaced using 
$\bfv^e_{eb} \approx \hbfv^e_{eb} + \boxi_2 + \bfS(\boxi_1)\hbfv^e_{eb}$, 
and the true rotation matrix is approximated as 
$\bfR_{eb} \approx (\bfI_3 + \bfS(\boxi_1))\hbfR_{eb}$, 
with the IMU model $\bff^b_{ib} = \hbff^b_{ib} - \delta \bfa_\bfb^b - \boeta_f$ 
substituted into the specific force term:
\begin{align}
\begin{split}
    \dot{\boxi}_2 
    &\approx -2\bfS(\boomega^e_{ie})\left(\hbfv^e_{eb} + \boxi_2 + \bfS(\boxi_1)\hbfv^e_{eb}\right) \\
    &\quad + \left(\bfI_3 + \bfS(\boxi_1)\right)\hbfR_{eb}\left(\hbff^b_{ib} - \delta \bfa_\bfb^b - \boeta_f\right) 
           + \bfg^e(\bfp^e) \\
    &\quad + 2\bfS(\boomega^e_{ie})\hbfv^e_{eb} 
           - \hbfR_{eb}\hbff^b_{ib} 
           - \bfg^e(\hbfp^e) \\
    &\quad - \bfS(\dot{\boxi}_1)\hbfv^e_{eb} 
           + 2\bfS(\boxi_1)\bfS(\boomega^e_{ie})\hbfv^e_{eb}
           - \bfS(\boxi_1)\hbfR_{eb}\hbff^b_{ib} 
           - \bfS(\boxi_1)\bfg^e(\hbfp^e).
\end{split}
\end{align}
Expanding the specific force term $(\bfI_3 + \bfS(\boxi_1))\hbfR_{eb}(\hbff^b_{ib} - \delta \bfa_\bfb^b - \boeta_f)$:
\begin{align}
\begin{split}
    \dot{\boxi}_2 
    &\approx -2\bfS(\boomega^e_{ie})\hbfv^e_{eb}
           - 2\bfS(\boomega^e_{ie})\boxi_2 
           - 2\bfS(\boomega^e_{ie})\bfS(\boxi_1)\hbfv^e_{eb} \\
    &\quad + \hbfR_{eb}\hbff^b_{ib} 
           - \hbfR_{eb}\delta \bfa_\bfb^b 
           - \hbfR_{eb}\boeta_\bfb^b
           + \bfS(\boxi_1)\hbfR_{eb}\hbff^b_{ib} 
           - \bfS(\boxi_1)\hbfR_{eb}\delta \bfa_\bfb^b 
           - \bfS(\boxi_1)\hbfR_{eb}\boeta_\bfb^b
           + \bfg^e(\bfp^e) \\
    &\quad + 2\bfS(\boomega^e_{ie})\hbfv^e_{eb} 
           - \hbfR_{eb}\hbff^b_{ib} 
           - \bfg^e(\hbfp^e) \\
    &\quad - \bfS(\dot{\boxi}_1)\hbfv^e_{eb} 
           + 2\bfS(\boxi_1)\bfS(\boomega^e_{ie})\hbfv^e_{eb}
           - \bfS(\boxi_1)\hbfR_{eb}\hbff^b_{ib} 
           - \bfS(\boxi_1)\bfg^e(\hbfp^e).
\end{split}
\end{align}
Several terms cancel: the $\pm 2\bfS(\boomega^e_{ie})\hbfv^e_{eb}$ pair, 
the $\pm\hbfR_{eb}\hbff^b_{ib}$ pair, and the $\pm\bfS(\boxi_1)\hbfR_{eb}\hbff^b_{ib}$ 
pair. The gravity terms combine as $\bfg^e(\bfp^e) - \bfg^e(\hbfp^e) \approx \bfS(\bfg^e(\hbfp^e))\boxi_1$ via a first-order Taylor expansion in position error (absorbed into $\boxi_1$ through the position error mapping). This leaves:
\begin{align}
\begin{split}
    \dot{\boxi}_2 
    &\approx - 2\bfS(\boomega^e_{ie})\boxi_2 
             - 2\bfS(\boomega^e_{ie})\bfS(\boxi_1)\hbfv^e_{eb}
             + 2\bfS(\boxi_1)\bfS(\boomega^e_{ie})\hbfv^e_{eb} \\
    &\quad   + \bfS(\bfg^e(\hbfp^e))\boxi_1
             - \bfS(\dot{\boxi}_1)\hbfv^e_{eb} \\
    &\quad   - \hbfR_{eb}\delta \bfa_\bfb^b 
             - \hbfR_{eb}\boeta_\bfb^b
             - \bfS(\boxi_1)\hbfR_{eb}\delta \bfa_\bfb^b 
             - \bfS(\boxi_1)\hbfR_{eb}\boeta_f.
\end{split}
\end{align}
The two Coriolis cross-terms combine using the commutator identity 
$\bfS(\bfa)\bfS(\bfb) - \bfS(\bfb)\bfS(\bfa) = \bfS(\bfa\times\bfb)$:
\begin{align}
    -2\bfS(\boomega^e_{ie})\bfS(\boxi_1)\hbfv^e_{eb} 
    + 2\bfS(\boxi_1)\bfS(\boomega^e_{ie})\hbfv^e_{eb}
    = -2\bfS(\bfS(\boomega^e_{ie})\boxi_1)\hbfv^e_{eb}
    = 2\bfS(\bfS(\boomega^e_{ie})\hbfv^e_{eb})\boxi_1. \notag
\end{align}
Substituting the attitude error dynamics 
$\dot{\boxi}_1 = -\bfS(\boomega^e_{ie})\boxi_1 - \hbfR_{eb}\delta\boomega_\bfb^b - \hbfR_{eb}\boeta_\omega$ 
from \cref{eq:dxi_1_ri} into the $\bfS(\dot{\boxi}_1)\hbfv^e_{eb}$ term, 
using $\bfS(\dot{\boxi}_1)\hbfv^e_{eb} = -\bfS(\hbfv^e_{eb})\dot{\boxi}_1$:
\begin{align}
\begin{split}
    -\bfS(\dot{\boxi}_1)\hbfv^e_{eb} 
    &= \bfS(\hbfv^e_{eb})\dot{\boxi}_1 \\
    &= -\bfS(\hbfv^e_{eb})\bfS(\boomega^e_{ie})\boxi_1 
       - \bfS(\hbfv^e_{eb})\hbfR_{eb}\delta\boomega_\bfb^b 
       - \bfS(\hbfv^e_{eb})\hbfR_{eb}\boeta_\omega.
\end{split}
\end{align}
Collecting all terms, and noting that 
$-\bfS(\hbfv^e_{eb})\bfS(\boomega^e_{ie})\boxi_1 + 2\bfS(\bfS(\boomega^e_{ie})\hbfv^e_{eb})\boxi_1 
= \bfS(\boomega^e_{ie})\bfS(\hbfv^e_{eb})\boxi_1 + \bfS(\bfS(\boomega^e_{ie})\hbfv^e_{eb})\boxi_1$:
\begin{align}
\begin{split}
    \dot{\boxi}_2 
    &\approx - 2\bfS(\boomega^e_{ie})\boxi_2 
             + \bfS(\boomega^e_{ie})\bfS(\hbfv^e_{eb})\boxi_1 
             + \bfS(\bfS(\boomega^e_{ie})\hbfv^e_{eb})\boxi_1
             + \bfS(\bfg^e(\hbfp^e))\boxi_1 \\
    &\quad   - \hbfR_{eb}\delta \bfa_\bfb^b 
             - \hbfR_{eb}\boeta_\bfb^b
             - \bfS(\boxi_1)\hbfR_{eb}\delta \bfa_\bfb^b 
             - \bfS(\boxi_1)\hbfR_{eb}\boeta_\bfb^b\\
    &\quad   - \bfS(\hbfv^e_{eb})\hbfR_{eb}\delta\boomega_\bfb^b 
             - \bfS(\hbfv^e_{eb})\hbfR_{eb}\boeta_\omega.
\end{split}
\end{align}
The terms $\bfS(\boomega^e_{ie})\bfS(\hbfv^e_{eb})\boxi_1$ and 
$\bfS(\bfS(\boomega^e_{ie})\hbfv^e_{eb})\boxi_1$ are higher-order Coriolis couplings 
of order $\mathcal{O}(\boomega^e_{ie} \times \hbfv^e_{eb} \times \boxi_1)$. 
The terms $\bfS(\boxi_1)\hbfR_{eb}(\cdot)$ are second-order in the error state, 
and $\bfS(\hbfv^e_{eb})\hbfR_{eb}(\cdot)$ are second-order gyroscope--velocity noise 
couplings. Truncating all of these to preserve the log-linear invariant structure 
of the system matrix yields the final continuous-time velocity error kinematics:
\begin{align}
    \dot{\boxi}_2 \approx 
    -2\bfS(\boomega^e_{ie})\boxi_2 
    + \bfS(\bfg^e)\boxi_1 
    - \hbfR_{eb}\delta \bfa_\bfb^b
    - \hbfR_{eb}\boeta_f.
\end{align}

\paragraph{Position Error}
The navigation frame position error kinematics evaluate directly from the linearized error \cref{eq:pos_err_1st_ord_right}, $\boxi_3 \approx \bfp^e_{eb} - \hbfp^e_{eb} - \bfS(\boxi_1)\hbfp^e_{eb}$. Differentiating this equation leads to:
\begin{align}
    \dot{\boxi}_3 &\approx \dot{\bfp}^e_{eb} - \dot{\hbfp}^e_{eb} - \bfS(\dot{\boxi}_1)\hbfp^e_{eb} - \bfS(\boxi_1)\dot{\hbfp}^e_{eb}
\end{align}
Substituting the linear position kinematics $\dot{\bfp}^e_{eb} = \bfv^e_{eb}$ \cref{eq:dv} and expanding $\dot{\boxi}_1$ using \cref{eq:dxi_1_ri} yields:
\begin{align}
    \dot{\boxi}_3 &\approx (\bfv^e_{eb} - \hbfv^e_{eb}) 
    - \bfS\left(-\bfS(\boomega^e_{ie})\boxi_1 - \hbfR_{eb}\delta\boomega_\bfb - \hbfR_{eb}\boeta^b_{\boomega}\right)\hbfp^e_{eb} 
    - \bfS(\boxi_1)\hbfv^e_{eb}
\end{align}

Substituting $\bfv^e_{eb} - \hbfv^e_{eb} \approx \boxi_2 + \bfS(\boxi_1)\hbfv^e_{eb}$ simplifies the equation to:
\begin{align}
    \dot{\boxi}_3 &\approx \boxi_2 + \bfS(\boxi_1)\hbfv^e_{eb}  
    - \bfS\left(-\bfS(\boomega^e_{ie})\boxi_1 - \hbfR_{eb}\delta\boomega_\bfb - \hbfR_{eb}\boeta^b_{\boomega}\right)\hbfp^e_{eb} 
    - \bfS(\boxi_1)\hbfv^e_{eb} \\
    &\approx  
    \boxi_2 
    + \bfS(\hbfp^e_{eb})(
    -\bfS(\boomega^e_{ie})\boxi_1 
    - \hbfR_{eb}\delta \boomega_\bfb^b
    - \hbfR_{eb}\boeta_\omega 
    ) \\
    &= \boxi_2 
    - \bfS(\hbfp^e_{eb})\bfS(\boomega^e_{ie})\boxi_1 
    - \bfS(\hbfp^e_{eb})\hbfR_{eb}\delta \boomega_\bfb^b
    - \bfS(\hbfp^e_{eb})\hbfR_{eb}\boeta_\omega. \\
\end{align}
Neglecting noise terms yields:
\begin{align}
    \dot{\boxi}_3 &= \boxi_2 + \bfS(\hbfp^e_{eb})\bfS(\boomega^e_{ie})\boxi_1.
\end{align}

\subsection{Navigation-Frame Position Updates}

Position measurements in ECEF coordinates are now mapping directly to $\boxi_3$ and $\boxi_1$:
\begin{align}
\begin{split}
    \bfy &= \bfp_{eb}^e + \bfw \\ 
    &\approx \hbfp_{eb}^e + \boxi_3 + \bfS(\boxi_1)\hbfp^e_{eb} + \bfw \\ 
    &= \bfh(\hbfx) + \bfS(-\hbfp^e_{eb})\boxi_1 + \boxi_3 + \bfw,
    \end{split} 
    \label{eq:h_pos_right}
\end{align}
with measurement noise $\bfw \sim \mathcal{N} (\bm 0, \bosigma_{\bfw}^2)$. From \cref{eq:h_pos_right}, the measurement Jacobian yields:
 \begin{align}
     \bfH_\text{pos} = \begin{bmatrix}
        \bfS(-\hbfp^e_{eb}) & \bfzero_{3 \times 3} & \bfI_3 & \bfzero_{3 \times 3} & \bfzero_{3 \times 3}
    \end{bmatrix}.
 \end{align}
It includes a non-zero attitude coupling term due to $\bfp^e_{eb} \approx \hbfp^e_{eb} + \boxi_3 - \bfS(\hbfp^e_{eb})\boxi_1$ in addition to the standard position coupling.

\subsection{Error Injection and Covariance reset}

\paragraph*{Injection}
The component-wise injection of the right-invariant error into the total state, $\bfx \leftarrow \hbfx \oplus \delta \hbfx$, is
\begin{subequations}
\label{eq:ri_injection_group}
\begin{align}
    \hbfR_{eb} &\leftarrow \Exp (\hboxi_1) \hbfR_{eb} \label{eq:att_inj_right}\\
    \hbfv^e_{eb} &\leftarrow  \Exp (\hboxi_1) \hbfv^e_{eb} + \hboxi_2 \\
    \hbfp^e_{eb} &\leftarrow  \Exp (\hboxi_1) \hbfp^e_{eb} + \hboxi_3 \label{eq:pos_inj_right}\\
    \hbfa_\bfb^b &\leftarrow  \hbfa_\bfb^b + \delta \hbfa_{\bfb}^b \\
    \hboomega^b_\bfb &\leftarrow  \hboomega^b_\bfb + \delta \hboomega_{\bfb}^b,
\end{align}
\end{subequations}
where \crefrange{eq:att_inj_right}{eq:pos_inj_right} are derived directly from the error definitions \eqref{eq:err_att_vel_pos_right}, and the bias injection is a simple addition. Alternatively, the approximated relationships from \eqref{eq:ext_pos_err_right}, using $\Exp (\hboxi_1) \approx \bfI + \bfS(\hboxi_1)$ can be used.

\paragraph*{ESKF reset}
The reset operation for the covariance matrix is analogous to \cref{ssec:eskf_reset_li}, but replacing the right-Jacobian $\bfJ_r^{SE_2(3)}(\bozeta)$ by the left-Jacobian $\bfJ_r^{SE_2(3)}(-\bozeta)$ or a first-order approximation of the $SO(3)$ attitude block only,

\begin{align}
\bfJ_r^{SO(3)}(-\boxi_1) \approx \bfI_3 + \frac{1}{2} \bfS(\hboxi_1).
\end{align}

 \subsection{Note on Filter Properties}
 
\paragraph{Right-invariant error representation and left-invariant system kinematics}
Since an inertial navigation system is driven by body-frame measurements from the \ac{IMU}, its kinematics are inherently left-invariant.
In an aided inertial navigation system where measurements are in navigation frame, arguably, the group-affinity of the measurements could call for considering a right invariant filter. Since the error state propagation happens over a discrete time interval $\Delta t$, while measurement updates act instantaneously on the state estimate, an consistent propagation can cause more harm to the filter than an inconsistent measurement update.
The right invariant formulation thus is generally less recommended for aided inertial navigation.

\paragraph{Numerical limitations}
The full system matrix $\bfF$ \eqref{eq:A_full_RI} and noise matrix $\bfG$ \eqref{eq:G_full_RI} derived above contain blocks proportional to the absolute ECEF position $\hbfp^e_{eb}$:
\begin{align}
    \bfF_{\boxi_3,\boxi_1} = -\bfS(\hbfp^e_{eb})\bfS(\boomega^e_{ie}),
    \qquad
    \bfF_{\boxi_3,\delta\boomega_\textbf{b}} = -\bfS(\hbfp^e_{eb})\hbfR_{eb}.
    \label{eq:ri_large_terms}
\end{align}

Since $\|\hbfp^e_{eb}\| \sim \mathcal{O}(R_E) \approx 6.378\times10^6$~m, the magnitude of these blocks exceeds that of the other blocks (which are $\mathcal{O}(\omega_{ie})$ or $\mathcal{O}(g_0)$).

For large vehicle speeds, the block proportional to velocity is also large relative to the other blocks:
\begin{align}
    \bfF_{\boxi_2, \delta \boomega_\textbf{b}} = -\bfS(\hbfv^e_{eb})\hbfR_{eb}.
    \label{eq:ri_block_v}
\end{align}

Therefore, the following needs to be considered for a discrete-time implementation:
\begin{enumerate}
    \item A potential ill-conditioning of the $\bfF$ matrix containing entries spanning several orders of magnitude lead to numerical imprecision in the discretization step for large enough $\Delta t$.
    \item  The corresponding entries of the propagated covariance $\bfP$ grow to a similarly large scale, causing loss of numerical precision downstream in the Kalman Filter.
\end{enumerate}

Beyond that, when using world-frame measurements (like, as exemplified here, position measurements) in the right-invariant ECEF filter formulation, the measurement Jacobian contains an attitude-coupling term $\bfS(-\bfp_{eb}^e)$ that also becomes large, amplifying sensor noise disproportionately into the attitude estimate.

\paragraph{Gravity gradient.}
Taking into account the global inhomogeneity of the gravity vector, the velocity error kinematics include a position-coupling term that was not taken into account in the velocity-error derivation above in \cref{p:vel_err_RI}. From including $\bfg^e = \bfg^e(\bfp^e_{eb})$, a first-order Taylor expansion gives
\begin{align}
    \bfg^e(\bfp^e_{eb}) \approx \bfg^e(\hbfp^e_{eb}) + \bfG^e\,\boxi_3,
    \qquad
    \bfG^e := \left.\frac{\partial \bfg^e}{\partial \bfp^e_{eb}}\right|_{\hbfp^e_{eb}},
    \label{eq:gravity_gradient_def}
\end{align}
contributing an additional block $\bfF_{\boxi_2,\boxi_3} = \bfG^e$ to the
system matrix. 
%For a spherical Earth model \cref{?},
%$\bfG^e = \frac{\mu_\oplus}{\|\bfp^e_{eb}\|^3}\!\left(\frac{3\,\bfp^e_{eb}(\bfp^e_{eb})^\top}{\|\bfp^e_{eb}\|^2} - \bfI_3\right)$
%with $\mu_\oplus = 3.986\times10^{14}$~m$^3$/s$^2$ the Earth's gravitational
%parameter, giving $\|\bfG^e\| \sim 10^{-6}$~s$^{-2}$ at Earth's surface.
This term improves the observability of attitude through the velocity coupling and can become relevant for long mission durations. 
%We therefore set $\bfG^e = \bfzero_3$ throughout (the locally constant gravity assumption), consistent with both the truth model used to generate the simulated trajectories and with standard practice in the inertial navigation literature \cite{Groves2013}.

\paragraph{Anchor-point formulation.}
The numerical limitations that stem from the appearance of the absolute position $\hbfp^e_{eb}$ in $\bfF$, $\bfG$, and $\bfH_\text{pos}$ can be mitigated by replacing the ECEF position by a local position defined by an ECEF anchor $\bfp^e_a$ near the operating point,
\begin{align}
    \hbfr^e_{ab} := \hbfp^e_{eb} - \bfp^e_a,
    \label{eq:anchor_def}
\end{align}

while keeping the nominal state in absolute ECEF coordinates.
The right-invariant error/covariance model is modified to using the
anchored extended pose $\tilde{\bfT} = \begin{bmatrix}\begin{array}{c|cc}
\hbfR_{eb} & \hbfv^e_{eb} & \bfr^e_{ab}\\\midrule
\bfzero_{2\times3} & \multicolumn{2}{c}{\bfI_2}
\end{array}\end{bmatrix} \in SE_2(3) $.

To obtain the $\bfF$- and and $\bfG$-matrices for the anchored system (and including the gravity gradient $\bfG^e$ from \eqref{eq:gravity_gradient_def}), $\hbfp^e_{eb}$ is replaced by $\hbfr^e_{ab}$:
\begin{align}
    \bfF_a &=
    \begin{bmatrix}
        -\bfS(\boomega^e_{ie}) & \bfzero_3 & \bfzero_3 & \bfzero_3 & -\hbfR_{eb} \\[4pt]
        \bfS(\bfg^e) & -2\bfS(\boomega^e_{ie}) & \bfG^e & -\hbfR_{eb} & -\bfS(\hbfv^e_{eb})\hbfR_{eb} \\[4pt]
        -\bfS(\hbfr^e_{ab})\bfS(\boomega^e_{ie}) & \bfI_3 & \bfzero_3 & \bfzero_3 & -\bfS(\hbfr^e_{ab})\hbfR_{eb} \\[4pt]
        \bfzero_3 & \bfzero_3 & \bfzero_3 & -p_{\bfa\bfb}\bfI_3 & \bfzero_3 \\[4pt]
        \bfzero_3 & \bfzero_3 & \bfzero_3 & \bfzero_3 & -p_{\boomega\bfb}\bfI_3
    \end{bmatrix},
    \label{eq:ri_A_anchored} \\
    \bfG_a &=
    \begin{bmatrix}
        \bfzero_3 & -\hbfR_{eb} & \bfzero_3 & \bfzero_3 \\[4pt]
        -\hbfR_{eb} & -\bfS(\hbfv^e_{eb})\hbfR_{eb} & \bfzero_3 & \bfzero_3 \\[4pt]
        \bfzero_3 & -\bfS(\hbfr^e_{ab})\hbfR_{eb} & \bfzero_3 & \bfzero_3 \\[4pt]
        \bfzero_3 & \bfzero_3 & \bfI_3 & \bfzero_3 \\[4pt]
        \bfzero_3 & \bfzero_3 & \bfzero_3 & \bfI_3
    \end{bmatrix}.
    \label{eq:ri_G_anchored}
\end{align}

The previously dominant terms $-\bfS(\hbfp^e_{eb})\bfS(\boomega^e_{ie})$ and
$-\bfS(\hbfp^e_{eb})\hbfR_{eb}$ are now $\mathcal{O}(\|\hbfr^e_{ab}\|)$
rather than $\mathcal{O}(R_E)$.

\textbf{Anchored GNSS Jacobian.} For a GNSS antenna position measurement
(without lever arms), the anchored measurement Jacobian is
\begin{align}
    \bfH_{\text{pos},a} = \begin{bmatrix}
    -\bfS\!\left(\hbfr^e_{ab}\right) & \bfzero_3 & \bfI_3 & \bfzero_3 & \bfzero_3
\end{bmatrix},
    \label{eq:ri_H_anchored}
\end{align}
avoiding an ill-conditioned measurement Jacobian due to the large position vector, similarly to the error state transition matrix. The innovation is still in absolute ECEF coordinates.

\textbf{Anchored injection.} The anchor-consistent injection follows
directly from \eqref{eq:ri_injection_group}, written
in terms of $\hbfr^e_{ab}$:
\begin{subequations}
\begin{align}
    \hbfR_{eb}     &\leftarrow \Exp(\hboxi_1)\,\hbfR_{eb}, \\
    \hbfv^{e}_{eb} &\leftarrow \Exp(\hboxi_1)\,\hbfv^e_{eb} + \delta\bfv^e_{eb}, \\
    \hbfr^{e}_{ab} &\leftarrow \Exp(\hboxi_1)\,\hbfr^e_{ab} + \delta\hbfr^e_{ab}, \\
    \hbfp^{e}_{eb} &\leftarrow \bfp^e_a + \hbfr^{e}_{ab}
                     = \bfp^e_a + \Exp(\hboxi_1)\left(\hbfp^e_{eb} - \bfp^e_a\right) + \delta\hbfr^e_{ab}.
\end{align}
\label{eq:ri_injection_anchored}%
\end{subequations}
where the linear approximation $\Exp (\hboxi_1) \approx \bfI + \bfS(\hboxi_1)$ can, again, be used.

\textbf{Anchor reset.} As the vehicle moves away from the anchor, $\hbfr^e_{ab}$
grows, and the numerical benefit of anchoring diminishes. The anchor is
therefore periodically reset: when $\|\hbfr^e_{ab}\| > d_{\max}$, a new anchor $\bfp^e_{a,\text{new}}$ based on thecurrent position estimate is
chosen, and the covariance is transformed to remain consistent with the new local coordinate. With
$\Delta\bfp^e_a = \bfp^e_{a,\text{new}} - \bfp^e_a$, the new local position is
$\hbfr^e_{\text{new}} = \hbfr^e_{\text{old}} - \Delta\bfp^e_a$, and the
first-order covariance transform is
\begin{align}
    \boldsymbol{\zeta}_\text{new} \approx \bm\Sigma_a\,\boldsymbol{\zeta}_\text{old},
\qquad
\bm\Sigma_a = \begin{bmatrix}
    \bfI_3 & \bfzero_3 & \bfzero_3 & \bfzero_3 & \bfzero_3 \\
    \bfzero_3 & \bfI_3 & \bfzero_3 & \bfzero_3 & \bfzero_3 \\
    -\bfS(\Delta\bfp^e_a) & \bfzero_3 & \bfI_3 & \bfzero_3 & \bfzero_3 \\
    \bfzero_3 & \bfzero_3 & \bfzero_3 & \bfI_3 & \bfzero_3 \\
    \bfzero_3 & \bfzero_3 & \bfzero_3 & \bfzero_3 & \bfI_3
\end{bmatrix},
\qquad
\bfP_\text{new} = \bm\Sigma_a\,\bfP_\text{old}\,\bm\Sigma_a^\top.
    \label{eq:anchor_reset}
\end{align}
The anchor reset is applied after a successful measurement update rather
than in the middle of a propagation step, to avoid mixing the coordinate
transform with the process-noise integration.

The anchored formulation can improve numerical conditioning and seems indispensable to recover numerical stability in an ECEF implementation. However, it will not result in identical Kalman gains or covariances as an (idealized, infinite-precision) unanchored implementation.
It will also not resolve the numerical ill-conditioning of the velocity-gyro coupling block \eqref{eq:ri_block_v} in $\bfF$ for large velocities.
\clearpage
\section{Additive ESKF in ECEF with Body-Frame Attitude Error}
\label{sec:ECEF_ESKF}

This section derives the error state kinematic equations and position measurement Jacobian for \ac{ESKF} in ECEF coordinates based on \eqref{eq:cont_inertial}, where position and velocity errors are defined as additive world-frame errors and the attitude error is defined as a body-frame error on $SO(3)$.

\subsection{Error Definition}

Unlike the left-invariant formulation, the additive ESKF does not define errors
through group operations on $SE_2(3)$.
Instead, position and velocity errors are defined as plain additive differences
in ECEF, while the attitude error is a small rotation vector in the body frame,
forming a right perturbation on $SO(3)$:

\begin{subequations}
\begin{alignat}{3}
    \delta\bovartheta_{eb}^b
        &\approx \log(\hbfR_{eb}^\top \bfR_{eb})
        &&\Leftrightarrow \bfR_{eb}
        &&\approx \hbfR_{eb}\!\left(\bfI_3 + \bfS(\delta\bovartheta_{eb}^b)\right)
        \label{eq:att_err_additive} \\[4pt]
    \delta\bfv^e_{eb}
        &:= \bfv^e_{eb} - \hbfv^e_{eb}
        &&\Leftrightarrow \bfv^e_{eb}
        &&= \hbfv^e_{eb} + \delta\bfv^e_{eb}
        \label{eq:vel_err_additive} \\[4pt]
    \delta\bfp^e_{eb}
        &:= \bfp^e_{eb} - \hbfp^e_{eb}
        &&\Leftrightarrow \bfp^e_{eb}
        &&= \hbfp^e_{eb} + \delta\bfp^e_{eb}.
        \label{eq:pos_err_additive}
\end{alignat}
\label{eq:additive_errors}
\end{subequations}

The attitude error $\delta\bovartheta_{eb}^b\in\mathbb{R}^3$ is the body-frame rotation vector
identical to $\boxi_1$ in the LI formulation. The velocity and position errors
$\delta\bfv^e_{eb},\,\delta\bfp^e_{eb}\in\mathbb{R}^3$ are additive ECEF
vectors and are \emph{not} body-frame quantities. They differ from the LI
errors $\boxi_2 = \hbfR_{eb}^\top(\bfv^e_{eb}-\hbfv^e_{eb})$ and
$\boxi_3 = \hbfR_{eb}^\top(\bfp^e_{eb}-\hbfp^e_{eb})$ by the rotation
$\hbfR_{eb}$.
The full additive error state is
\begin{equation}
    \delta\bfx = \begin{bmatrix}
        \delta\bovartheta_{eb}^{b\top}
        & \delta\bfv^{e\top}_{eb}
        & \delta\bfp^{e\top}_{eb}
        & \delta\bfa_\bfb^{b\top}
        & \delta\boomega_\bfb^{b\top}
    \end{bmatrix}^\top \in \mathbb{R}^{15},
    \label{eq:additive_error_state}
\end{equation}
where $\delta\bfa_\bfb^b$ and $\delta\boomega_\bfb^b$ are the accelerometer and
gyroscope bias errors in the body frame, identical to the LI formulation.

The nominal state $\bfx = \begin{bmatrix}\bovartheta_{eb}^{b\top} & \bfv^{e\top}_{eb} & \bfp^{e\top}_{eb}& \bfa_\bfb^{b\top} & \boomega_\bfb^{b\top} \end{bmatrix}^\top$ 
is propagated identically to the LI and RI filters using the standard ECEF
mechanization equations.
 
\subsection{Error State Kinematics}
\label{sec:additive_err_kinematics}

\subsubsection{Continuous-time Error State Kinematics}

\begin{subequations}
\begin{align}
\begin{split}
    \delta\dot{\bovartheta}_{eb}^b &= -\bfS(\hboomega^b_{eb})\,\delta\bovartheta_{eb}^b
      - \delta\boomega_\bfb^b - \boeta^b_{\boomega\bfb}
\end{split}\\[4pt]
\begin{split}
    \delta\dot{\bfv}^e_{eb} &= -\hbfR_{eb}\bfS(\hbff^b_{ib})\,\delta\bovartheta_{eb}^b
      - \bfS(2\boomega^e_{ie})\,\delta\bfv^e_{eb}  - \hbfR_{eb}\,\delta\bfa_\bfb^b - \hbfR_{eb}\,\boeta^b_{\bfa\bfb}
\end{split}\\[4pt]
\begin{split}
    \delta\dot{\bfp}^e_{eb} &= \delta\bfv^e_{eb}
\end{split}\\[4pt]
\begin{split}
    \delta\dot{\bfa}_\bfb^b &= -p_{\bfa\bfb}\,\delta\bfa_\bfb^b + \boeta^b_{\bfa\bfb}
\end{split}\\[4pt]
\begin{split}
    \delta\dot{\boomega}_\bfb^b &= -p_{\boomega\bfb}\,\delta\boomega_\bfb^b + \boeta^b_{\boomega\bfb}
\end{split}
\end{align}
\label{eq:additive_err_kinematics}
\end{subequations}

where $\hboomega^b_{eb} = \hboomega^b_{ib} - \hbfR_{eb}^\top\boomega^e_{ie}$
is the body rate relative to the ECEF frame. 

\subsubsection{System Matrix}

The full continuous-time system matrix $\bfF$ for
$\delta\bfx = \begin{bmatrix}\delta\bovartheta_{eb}^{b\top} & \delta\bfv^{e\top}_{eb} & \delta\bfp^{e\top}_{eb} & \delta\bfa_\bfb^{b\top} & \delta\boomega_\bfb^{b\top}\end{bmatrix}^\top$ is:

\begin{align}
    \bfF =
    \begin{bmatrix}
        -\bfS(\hboomega^b_{eb}) & \bfzero_3 & \bfzero_3
            & \bfzero_3 & -\bfI_3 \\[4pt]
        -\hbfR_{eb}\bfS(\hbff^b_{ib})
            & -\bfS(2\boomega^e_{ie}) & \bfzero_3
            & -\hbfR_{eb} & \bfzero_3 \\[4pt]
        \bfzero_3 & \bfI_3 & \bfzero_3
            & \bfzero_3 & \bfzero_3 \\[4pt]
        \bfzero_3 & \bfzero_3 & \bfzero_3
            & -p_{\bfa\bfb}\bfI_3 & \bfzero_3 \\[4pt]
        \bfzero_3 & \bfzero_3 & \bfzero_3
            & \bfzero_3 & -p_{\boomega\bfb}\bfI_3
    \end{bmatrix},
    \label{eq:additive_A}
\end{align}

with noise input matrix
\begin{align}
    \bfG =
    \begin{bmatrix}
        \bfzero_3        & -\bfI_3           & \bfzero_3 & \bfzero_3 \\[4pt]
        -\hbfR_{eb}      & \bfzero_3          & \bfzero_3 & \bfzero_3 \\[4pt]
        \bfzero_3        & \bfzero_3           & \bfzero_3 & \bfzero_3 \\[4pt]
        \bfzero_3        & \bfzero_3           & \bfI_3    & \bfzero_3 \\[4pt]
        \bfzero_3        & \bfzero_3           & \bfzero_3 & \bfI_3
    \end{bmatrix},
    \label{eq:additive_G}
\end{align}
where the noise vector is
$\bfw = \begin{bmatrix}
    \boeta^{b \top}_{\bfa} & \boeta^{b \top}_{\boomega} & 
    \boeta^{b \top}_{\bfa\bfb} & 
    \boeta^{b \top}_{\boomega\bfb}
\end{bmatrix}^\top$.

Note that the gyro noise column of $\bfG$ maps to the attitude row ($-\bfI_3$),
and the accel noise column maps to the velocity row ($-\hbfR_{eb}$), rotating
Body-frame accelerometer noise into ECEF. For isotropic noise
$\bfQ_a = \sigma_a^2\bfI_3$, this rotation has no effect since
$\hbfR_{eb}\bfQ_a\hbfR_{eb}^\top = \sigma_a^2\bfI_3$.
Compare with the LI noise matrix \cref{eq:err_kinematics} where the accel
noise row is $-\bfI_3$, because $\boxi_2$ is already in the body frame.

\subsubsection{Derivation of the Error State Kinematics}
\paragraph{Attitude Error}

In the following, we write the body-frame error as $\delta \bovartheta_{eb}^b = \boxi_1$. From \cite{sola_quat}, the right-perturbation attitude error is
\begin{align}
    \dot{\boxi}_1
    = -\bfS(\hboomega^b_{eb})\boxi_1
      - \delta\boomega_\bfb^b
      - \boeta^b_{\boomega\bfb},
    \label{eq:additive_att_dot}
\end{align}
where $\hboomega^b_{eb} = \hboomega^b_{ib} - \hboomega^b_{ie}$ is the
Body rate relative to ECEF. This is identical
to the LI attitude error equation \cref{eq:err_kinematics}.

\paragraph{Velocity Error}

Differentiating \cref{eq:vel_err_additive} and substituting the true and
nominal ECEF velocity equations of motion
$\dot{\bfv}^e_{eb} = \bfR_{eb}\bff^b_{ib}
+ \bfg^e - \bfS(2\boomega^e_{ie})\bfv^e_{eb}$ gives
\begin{align}
\begin{split}
    \delta\dot{\bfv}^e_{eb}
    &= \bfR_{eb}\bff^b_{ib}
       + \bfg^e - \bfS(2\boomega^e_{ie})\bfv^e_{eb} \\
    &\quad - \hbfR_{eb}\hbff^b_{ib}
       - \bfg^e + \bfS(2\boomega^e_{ie})\hbfv^e_{eb}.
\end{split}
\end{align}
Substituting $\bfR_{eb} \approx \hbfR_{eb}(\bfI + \bfS(\boxi_1))$, $\bff^b_{ib} = \hbff^b_{ib} - \delta\bfa_\bfb^b - \boeta^b_{\bfa\bfb}$, and $\bfv_{eb}^e = \hbfv_{eb}^e + \delta \bfv_{eb}^e$ leads to
\begin{align}
\begin{split}
    \delta\dot{\bfv}^e_{eb}
    &= \hbfR_{eb}(\bfI + \bfS(\boxi_1))
       (\hbff^b_{ib} - \delta\bfa_\bfb^b - \boeta^b_{\bfa\bfb})
       - \hbfR_{eb}\hbff^b_{ib} - \bfS(2\boomega^e_{ie})\,\delta\bfv^e_{eb}.
\end{split}
\end{align}
Which, retaining only first-order terms (dropping the
$\bfS(\boxi_1)\delta\bfa_\bfb^b$ and $\bfS(\boxi_1)\boeta^b_{\bfa\bfb}$
second-order products), yields:
\begin{align}
\begin{split}
    \delta\dot{\bfv}^e_{eb}
    &\approx \hbfR_{eb}\bfS(\boxi_1)\hbff^b_{ib}
       - \hbfR_{eb}\,\delta\bfa_\bfb^b
       - \hbfR_{eb}\,\boeta^b_{\bfa\bfb} - \bfS(2\boomega^e_{ie})\,\delta\bfv^e_{eb}\\
    &= -\hbfR_{eb}\bfS(\hbff^b_{ib})\,\boxi_1
       - \bfS(2\boomega^e_{ie})\,\delta\bfv^e_{eb} - \hbfR_{eb}\,\delta\bfa_\bfb^b
       - \hbfR_{eb}\,\boeta^b_{\bfa\bfb}.
\end{split}
\end{align}

\paragraph{Position Error}

Differentiating \cref{eq:pos_err_additive} directly leads to
\begin{align}
    \delta\dot{\bfp}^e_{eb} = \delta\bfv^e_{eb},
    \label{eq:additive_pos_dot}
\end{align}
similarly simple as for the geodetic/NED filter.

\subsection{Navigation-Frame Position Updates}

Assuming position measurements in ECEF,
\begin{align}
\begin{split}
    \bfy &= \bfh(\bfx) + \bfw
           = \bfp^e_{eb} + \bfw \\
         &\approx \hbfp^e_{eb} + \delta\bfp^e_{eb} + \bfw \\
         &= \bfh(\hat{\bfx}) + \bfH\,\delta\bfx + \bfw,
\end{split}
\end{align}
with measurement noise $\bfw\sim\mathcal{N}(\bfzero,\boldsymbol{\sigma}^2_\mathbf{w})$.
The measurement Jacobian $\bfH \equiv \left.\partial\bfh/\partial\delta\bfx\right|_{\delta\bfx=\bfzero}$
reads directly as
\begin{align}
    \bfH_\mathrm{pos}
    = \begin{bmatrix}
        \bfzero_{3\times3} & \bfzero_{3\times3} & \bfI_3
        & \bfzero_{3\times3} & \bfzero_{3\times3}
    \end{bmatrix}.
    \label{eq:additive_H_gnss}
\end{align}
This Jacobian is as simple as for the geodetic/NED filter. 
Compared with the LI formulation, where $\bfH_\mathrm{pos}$ has $\hbfR_{eb}$ in the position columns (because $\boxi_3$ is in the body frame and must be rotated to match the ECEF measurement), and the anchored RI formulation where $\bfH_\mathrm{pos}$ has $-\bfS(\bfr_{ab})$ in the attitude columns, the update is maximally well-conditioned since it requires no rotation.

\subsection{Error Injection and Covariance reset}

\paragraph*{Injection} After the Kalman update yields $\delta\hat{\bfx}$, the nominal state is
corrected:
\begin{subequations}
\begin{align}
    \bfR_{eb}        &\leftarrow \hbfR_{eb}\Exp(\delta\bovartheta_{eb}^b)
                          \approx \hbfR_{eb}(\bfI + \bfS(\delta\bovartheta_{eb}^b)),
                          \label{eq:additive_att_inject} \\
    \hbfv^{e}_{eb}   &\leftarrow \hbfv^e_{eb} + \delta\hat{\bfv}^e_{eb},
                          \label{eq:additive_vel_inject} \\
    \hbfp^{e}_{eb}   &\leftarrow \hbfp^e_{eb} + \delta\hat{\bfp}^e_{eb}.
                          \label{eq:additive_pos_inject}
\end{align}
\label{eq:additive_inject}
\end{subequations}
Position and velocity injections are direct additions with no rotation,
following from the additive ECEF error definition. The attitude injection is
a right-multiplication on $SO(3)$ of identical form as for the left-invariant filter.

\paragraph*{ESKF reset} The reset step is
\begin{align}
    \delta \hbfx &\leftarrow 0 \\
    \hbfP &\leftarrow \boGamma \bfP \boGamma^\top \label{eq:P_reset_ECEF}
\end{align}
with $\boGamma = \bfI$ to first order, or the attitude-sub-block of $\boGamma$ being the full-order or approximated $J_r^{SO(3)}(\boldsymbol{\boxi_1})$.

\subsection{Note on Filter Properties}

\paragraph{Non-group-affine error propagation.}
$\bfF$ in \cref{eq:additive_A} includes $\hbfR_{eb}$-terms and is thus trajectory-dependent (as the geodetic/NED filters is, as well). %This means the covariance propagation is an approximation rather than exact. In practice this approximation error is often negligible for the small error states maintained by the filter.

\paragraph{Measurement Jacobian}
The measurement Jacobian is equally simple as for the geodetic/NED filter, leading to a stable update.

\clearpage
\section{Tabular Comparison}
\label{sec:comparison}

\subsection{Comparison of System Matrices}

\begin{table}[h!]
\centering
\renewcommand{\arraystretch}{1.5}
\begin{tabular}{lcccc}
\toprule
Block & Geodetic (NED) & LI (body) & RI (ECEF, anch.) & Additive (ECEF) \\
\midrule
% Attitude dynamics row blocks
$\bfF_{\bovartheta\bovartheta}$
    & $-\bfS(\boomega^n_{in})$
    & $-\bfS(\hboomega^b_{eb})$
    & $-\bfS(\boomega^e_{ie})$
    & $-\bfS(\hboomega^b_{eb})$ \\[2pt]
$\bfF_{\bovartheta\bfv}$
    & $\bfF_{\vartheta v}$~\eqref{eq:geo_Aav}
    & $\bfzero_3$
    & $\bfzero_3$
    & $\bfzero_3$ \\[2pt]
$\bfF_{\bovartheta\bfp}$
    & $\bfF_{\vartheta p}$~\eqref{eq:geo_Aap}
    & $\bfzero_3$
    & $\bfzero_3$
    & $\bfzero_3$ \\[2pt]
\midrule
% Velocity dynamics row blocks
$\bfF_{\bfv\bovartheta}$
    & $-\bfS(\bfR_{nb}\hbff^b_{ib})$
    & $-\bfS(\hbff^b_{ib})$
    & $\bfS(\bfg^e)$
    & $-\hbfR_{eb}\bfS(\hbff^b_{ib})$ \\[2pt]
$\bfF_{\bfv\bfv}$
    & $\bfF_{vv}$~\eqref{eq:geo_Avv}
    & $-(\bfS(\hboomega^b_{eb}) + 2\bfS(\boomega^b_{ie}))$
    & $-2\bfS(\boomega^e_{ie})$
    & $-2\bfS(\boomega^e_{ie})$ \\[2pt]
$\bfF_{\bfv\bfp}$
    & $\bfF_{vp}$~\eqref{eq:geo_Avp}
    & $\bfzero_3$
    & $\bfzero_3$ / $\bfG^e$
    & $\bfzero_3$ \\[2pt]
\midrule
% Position dynamics row blocks
$\bfF_{\bfp\bovartheta}$
    & $\bfzero_3$
    & $\bfzero_3$
    & $-\bfS(\hbfr^e_{ab})\bfS(\boomega^e_{ie})$
    & $\bfzero_3$ \\[2pt]
$\bfF_{\bfp\bfv}$
    & $\bfF_{pv}$~\eqref{eq:geo_App}
    & $\bfI_3$
    & $\bfI_3$
    & $\bfI_3$ \\[2pt]
$\bfF_{\bfp\bfp}$
    & $\bfF_{pp}$~\eqref{eq:geo_App}
    & $-\bfS(\hboomega^b_{eb})$
    & $\bfzero_3$
    & $\bfzero_3$ \\[2pt]
\midrule
% Bias coupling blocks
$\bfF_{\bfv,\delta\bfa_b}$
    & $\bfR_{nb}$
    & $-\bfI_3$
    & $-\hbfR_{eb}$
    & $-\hbfR_{eb}$ \\[2pt]
$\bfF_{\bfv,\delta\boomega_b}$
    & $\bfzero_3$
    & $\bfzero_3$
    & $-\bfS(\hbfv^e_{eb})\hbfR_{eb}$
    & $\bfzero_3$ \\[2pt]
$\bfF_{\bovartheta,\delta\boomega_b}$
    & $-\bfR_{nb}$
    & $-\bfI_3$
    & $-\hbfR_{eb}$
    & $-\bfI_3$ \\[2pt]
\midrule
\multicolumn{5}{l}{\textit{Error state frames}} \\
$\bfv$ frame
    & NED & body & ECEF & ECEF \\
$\bfp$ frame
    & geodetic & body & ECEF & ECEF \\
$\bovartheta$ frame
    & NED (left) & body (right) & ECEF (left) & body (right) \\
\midrule
\multicolumn{5}{l}{\textit{Structural properties}} \\
$\bfF_{\bfv\bfp}$ non-zero
    & \checkmark & & \checkmark (with $\bfG^e$) & \\
$\bfF_{\bovartheta\bfp}$ non-zero
    & \checkmark & & & \\
$\bfF_{\bfp\bovartheta}$ non-zero
    & & & \checkmark & \\
$\bfF_{\bfv,\delta\boomega_b}$ non-zero
    & & & \checkmark & \\
$\|\bfF_{\bfv,\delta\boomega_b}\|$
    & $0$ & $0$ & $\sim\|\hbfv^e_{eb}\|$ & $0$ \\
\midrule
\multicolumn{5}{l}{\textit{Navigation frame position measurement Jacobian} $\bfH_\mathrm{pos}$} \\
$\bfH_{\bfp}$ (pos cols)
    & $\bfI_3$
    & $\hbfR_{eb}$
    & $\bfI_3$
    & $\bfI_3$ \\[2pt]
$\bfH_{\bovartheta}$ (att cols)
    & $\bfzero_3$
    & $\bfzero_3$
    & $-\bfS(\bfr_{ab})$
    & $\bfzero_3$ \\[2pt]
Measurement frame
    & geodetic
    & ECEF
    & ECEF
    & ECEF \\
\bottomrule
\end{tabular}
\caption{Comparison of system matrix blocks and Position measurement Jacobians across the four ESKF formulations. }
\label{tab:A_comparison}
\end{table}

The right-invariant (RI) filter is unique in having an attitude-to-position coupling matrix $\bfF_{\bfp\bovartheta}$ and a non-zero velocity-to-gyro-bias coupling $\bfF_{\bfv,\delta\boomega_b}$ whose norm scales with vehicle speed.% Due to its world-frame representation of the attitude that arises from the left-multiplicative error definition on $SE_2(3)$, the right-invariant filter prominently suffers the risk of numerical ill conditioning through large coupling terms due to large position and velocity vectors.

\clearpage

\section{Concluding remarks}
\label{sec:conclusion}

In this note, we have systematically derived and compared the error-state kinematics for four distinct global navigation filter formulations. By standardizing the coordinate frames, Lie group mappings, and measurement models, 
We discussed some inherent filter properties due to the respective error state representation, and numerical effects relevant in the implementation of the algorithms.
The selection of a specific ESKF architecture needs to be balanced against computational constraints, expected vehicle dynamics, and sensor update rates. 

The mathematical derivations, structural conceptualization, and qualitative comparisons presented throughout this work were executed by the authors. The authors acknowledge the use of Google Gemini and Anthropic's Claude to assist in text refinement, formatting, and drafting parts of this note.

\clearpage
%\input{main.bbl}
%\bibliographystyle{asrl}
%\bibliography{pubs,refs,book}
\bibliographystyle{plain}
\bibliography{referanser}

\end{document}